\documentclass{article} % For LaTeX2e
\usepackage[final]{colm2026_conference}

\usepackage[T1]{fontenc}
\usepackage{microtype}
\usepackage{url}
\usepackage{booktabs}
\usepackage{multirow}
\usepackage{amsmath}
\usepackage{amsfonts}
\usepackage{amssymb}
\usepackage{bm}
\usepackage{amsthm}
\usepackage{algorithm}
\usepackage{subcaption}
\usepackage{enumitem}
\usepackage{algorithmic}
\usepackage{makecell}
\usepackage{longtable}
\usepackage{colortbl}
\usepackage{listings}
\usepackage{pifont}
\usepackage{tabularx}
\usepackage{graphicx}
\usepackage{tcolorbox}
\usepackage[table]{xcolor}
\usepackage{wrapfig}
\usepackage{siunitx}
\usepackage{hyperref}

\usepackage{lineno}

\theoremstyle{remark}

\definecolor{darkblue}{rgb}{0, 0, 0.5}
\hypersetup{colorlinks=true, citecolor=darkblue, linkcolor=darkblue, urlcolor=darkblue}

\newcommand{\colortoprule}{\specialrule{\heavyrulewidth}{0pt}{0pt}}
\newcommand{\colormidrule}{\specialrule{\lightrulewidth}{0pt}{0pt}}
\newcommand{\colorbottomrule}{\specialrule{\heavyrulewidth}{0pt}{0pt}}

\title{\raisebox{-0.20\height}{\includegraphics[height=2.0em]{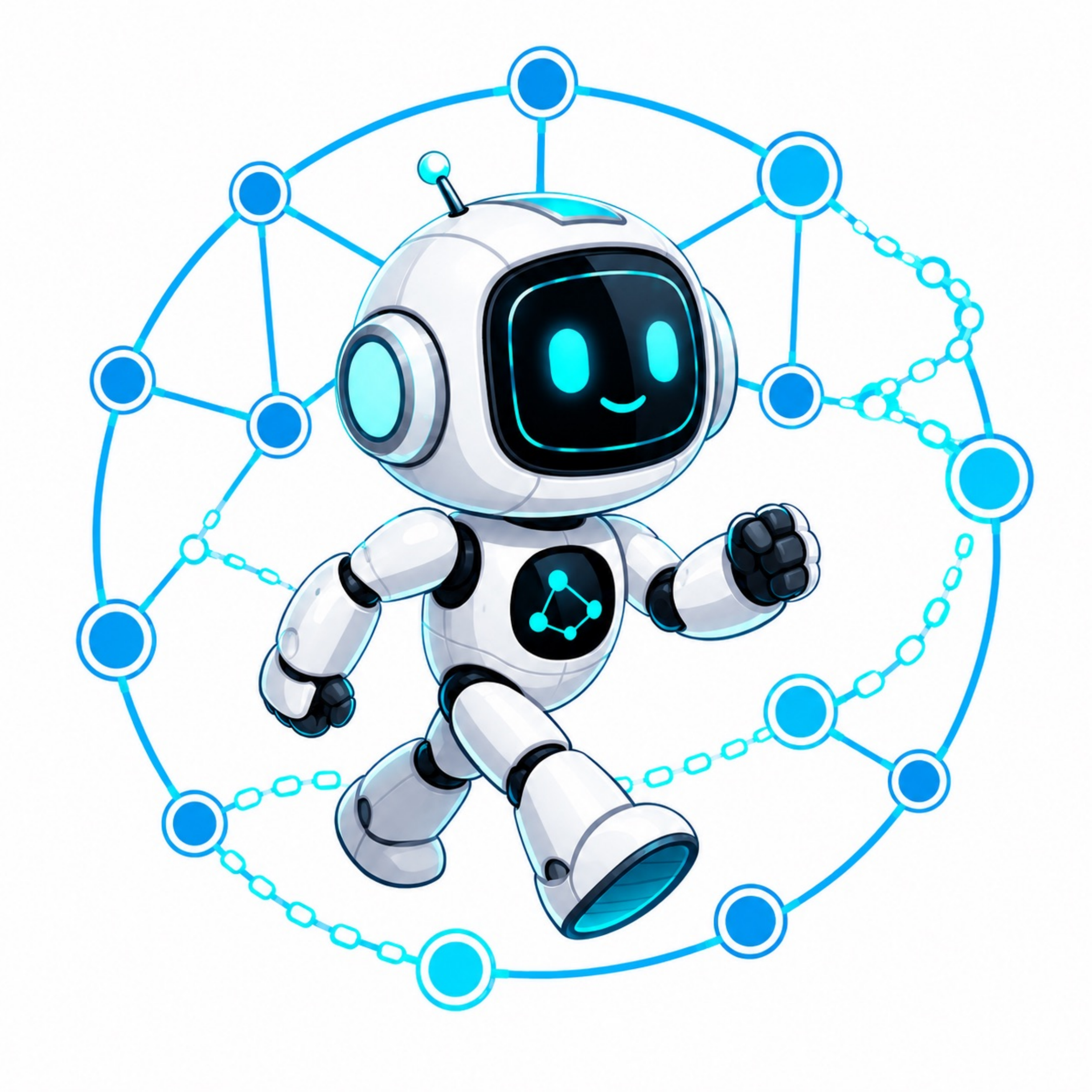}}~GraphWalker: Agentic Knowledge Graph Question Answering via Synthetic Trajectory Curriculum}

\author{Shuwen Xu$^{1\ast}$, Yao Xu$^{1,2\ast}$, Jiaxiang Liu$^{1,2}$, Chenhao Yuan$^1$, Wenshuo Peng$^3$, \\
\textbf{Jun Zhao}$^{1,2}$,
\textbf{Kang Liu}$^{1,2\dagger}$ \\
$^1$ School of Artificial Intelligence, University of Chinese Academy of Sciences \\
$^2$ The Key Laboratory of Cognition and Decision Intelligence for Complex Systems, \\ Institute of Automation, Chinese Academy of Sciences\\ 
$^3$ Department of Electronic Engineering, Tsinghua University \\ 
\texttt{xushuwen23@mails.ucas.ac.cn, \{yao.xu,jzhao,kliu\}@nlpr.ia.ac.cn}
}

\newcommand{\cmark}{\textcolor{green!60!black}{\ding{51}}\hspace{0.5em}}
\newcommand{\xmark}{\textcolor{red!80!black}{\ding{55}}\hspace{0.5em}}

\lstdefinestyle{sparql}{
  language=SQL,
  basicstyle=\ttfamily\small,
  keywordstyle=\color{blue}\bfseries,
  stringstyle=\color{red},
  commentstyle=\itshape\color{gray},
  showstringspaces=false,
  breaklines=true,
  frame=single,
  numbers=left,
  numberstyle=\tiny,
  captionpos=b,
}
\newcommand\blfootnote[1]{%
  \begingroup
  \renewcommand\thefootnote{}\footnote{#1}%
  \addtocounter{footnote}{-1}%
  \endgroup
}

\begin{document}

\ifcolmsubmission
\linenumbers
\fi

\maketitle
\blfootnote{$^\ast$ Equal contribution. $^\dagger$Corresponding author.}

\vspace{-3em}
\begin{abstract}
Agentic knowledge graph question answering (KGQA) requires an agent to iteratively interact with knowledge graphs (KGs), posing challenges in both training data scarcity and reasoning generalization. Specifically, existing approaches often restrict agent exploration: prompting-based methods lack autonomous navigation training, while current training pipelines usually confine reasoning to predefined trajectories. To this end, this paper proposes \textbf{\textit{GraphWalker}}, a novel agentic KGQA framework that addresses these challenges through \textit{Automated Trajectory Synthesis} and \textit{Stage-wise Fine-tuning}. GraphWalker adopts a two-stage SFT training paradigm: First, the agent is trained on structurally diverse trajectories synthesized from constrained random-walk paths, establishing a broad exploration prior over the KG; Second, the agent is further fine-tuned on a small set of expert trajectories to develop reflection and error recovery capabilities. Extensive experiments demonstrate that our stage-wise SFT paradigm unlocks a higher performance ceiling for a lightweight reinforcement learning (RL) stage, enabling GraphWalker to achieve state-of-the-art performance on CWQ and WebQSP. Additional results on GrailQA and our constructed GraphWalkerBench confirm that GraphWalker enhances generalization to out-of-distribution reasoning paths. The code is publicly available at \url{https://github.com/XuShuwenn/GraphWalker}
\end{abstract}

\section{Introduction}
\label{sec:intro}

Knowledge graph question answering (KGQA) aims to answer natural language questions by reasoning over large-scale structured Knowledge Graphs (KGs) \citep{jang2017kbqa}, where world knowledge is encoded as (subject, predicate, object) triples. Answering complex questions over KGs demands multi-hop reasoning, systematic path planning, and structural awareness of the underlying graph topology \citep{xiongenhancing}. The advent of Large Language Models (LLMs) \citep{brown2020language} has opened new opportunities for KGQA due to their strong language understanding and reasoning capabilities \citep{petroni2019language}.

\begin{figure}[h]
    \vspace{-10pt}
    \centering
    \includegraphics[width=\linewidth]{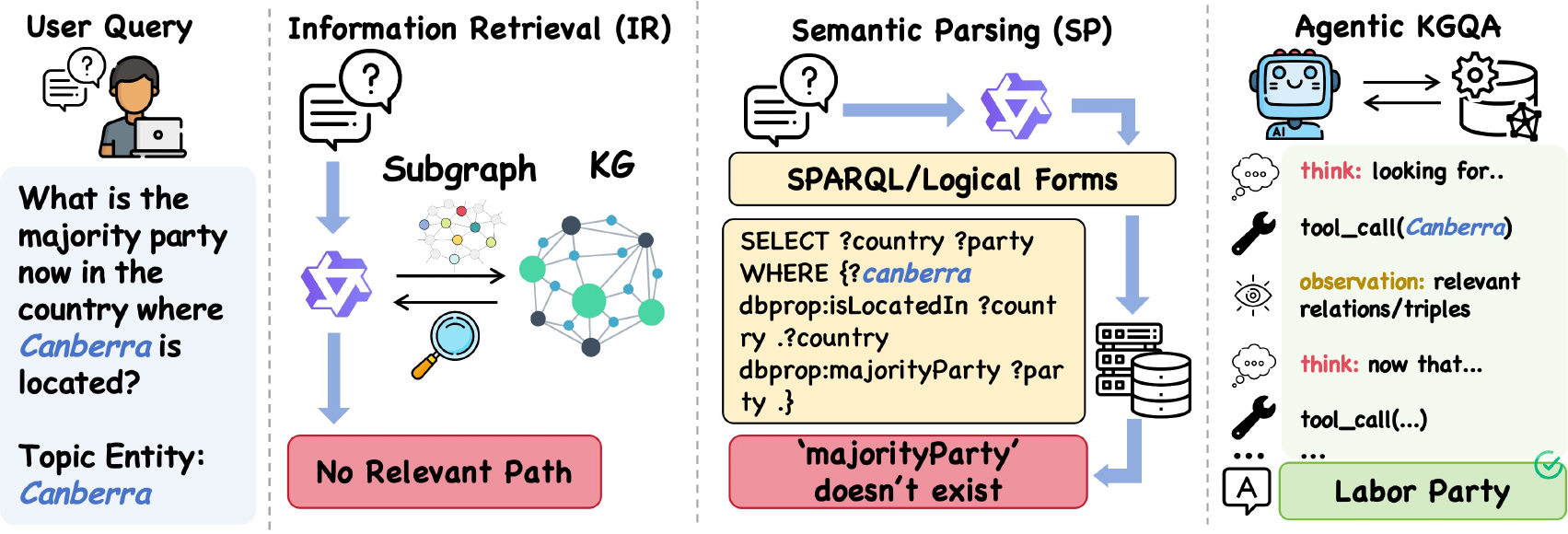}
    \caption{Comparison of three KGQA paradigms.}
    \label{fig:paradigm-comparison}
    \vspace{-10pt}
\end{figure}

As illustrated in Figure~\ref{fig:paradigm-comparison}, traditional KGQA methods follow two main paradigms:
\textbf{Information Retrieval (IR) methods} \citep{Zhang_2022, dong2023bridgingkbtextgapleveraging, li2024chainofknowledgegroundinglargelanguage}, which retrieve relevant subgraphs and extract answers from the gathered evidence, and \textbf{Semantic Parsing (SP) methods} \citep{berant2013semantic, yih2015semantic, li2023fewshotincontextlearningknowledge, Luo_2024}, which translate questions into executable logical forms (e.g., SPARQL) to query KGs directly. 
However, both paradigms confine reasoning within a static, pre-defined scope, which severely limits their \textit{generalization to unseen reasoning structures} in real-world KGs \citep{auer2007dbpedia, bollacker2008freebase, vrandecic2014wikidata, xiong2025deliberate}.

% To overcome this limitation, recent \textbf{Agentic KGQA methods} empower LLMs to iteratively interact with the global KG via successive tool calls, dynamically refining their reasoning based on real-time KG feedback \citep{yao2023reactsynergizingreasoningacting, sun2023think}.
% Despite this progress, as shown in Table~\ref{tab:method-comparison}, existing approaches remain limited in several aspects. 
% \textbf{Prompting-based methods}~\citep{sun2023think,xu2024generate} offer flexible reasoning but lack the domain-specific adaptation needed to reliably navigate noisy and complex KGs.
% \textbf{Training-based methods}, on the other hand, lack unified capabilities: 
% RoG \citep{luo2023reasoning} restricts reasoning to predefined workflows, lacking \textit{autonomous interaction}; 
% KG-Agent \citep{jiang2025kg} relies on imitation without \textit{RL optimization} to break the demonstration ceiling; 
% existing RL methods (e.g., KG-R1 \citep{song2025efficient}) often rely on pre-sampled subgraphs rather than the \textit{global KG}.
% Critically, recent studies have consistently shown that RL optimization requires a competent SFT prior: without sufficient exploration capacity, the agent cannot discover rewarding trajectories, rendering policy optimization ineffective \citep{Guo_2025, zhang2026asteragenticscalingtoolintegrated, xiong2026scaling}.
% This raises a central question: \textit{How can an agent build a robust exploration foundation on KGs to push the boundaries of its reasoning capacity?}

To overcome this limitation, recent \textbf{Agentic KGQA methods} allow LLMs to interact with KGs via successive tool calls, dynamically refining their reasoning based on real-time feedback \citep{yao2023reactsynergizingreasoningacting, sun2023think}.
However, as shown in Table~\ref{tab:method-comparison}, existing approaches still exhibit notable limitations.
Prompting-based and SFT-based methods \citep{sun2023think,xu2024generate,luo2023reasoning,jiang2025kg} often hit performance ceilings due to \textit{insufficient environmental adaptation}. 
Meanwhile, existing RL-based methods \citep{song2025efficient} suffer from a critical exploration bottleneck: they apply policy optimization on pre-extracted subgraphs without a comprehensive SFT prior. 
By confining their training to simplified environments, they severely restrict the agent's search space, hindering effective navigation of the vast, noisy \textit{global KG} during inference.
Most importantly, recent studies \citep{Guo_2025, zhang2026asteragenticscalingtoolintegrated, xiong2026scaling} consistently demonstrate that RL requires a structurally diverse SFT foundation; without a broad exploration prior, policy optimization on global KGs proves ineffective.
This raises a central question: \textit{\textbf{How can an agent build a robust exploration foundation to push the boundaries of its reasoning capacity?}}

\begin{table}[t]
  \centering
  \small
  \setlength{\tabcolsep}{5pt}
  \renewcommand{\arraystretch}{1.25}
  \begin{tabular}{l|l|ccc}
  \toprule
  \textbf{Type}
  & \textbf{Method}
  & \textbf{\shortstack{Global\\KG}}
  & \textbf{\shortstack{Autonomous\\Interaction}}
  & \textbf{\shortstack{RL \\ Optimization}} \\
  \midrule
  \multirow{2}{*}{Prompting}
  & ToG \citep{sun2023think}         & \cmark & \cmark & \xmark \\
  & GoG \citep{xu2024generate}       & \cmark & \cmark & \xmark \\
  \midrule
  \multirow{4}{*}{Training}
  & RoG \citep{luo2023reasoning}     & \xmark & \xmark & \xmark \\
  & KBQA-o1 \citep{luo2025kbqa}     & \cmark & \cmark & \xmark \\
  & KG-Agent \citep{jiang2025kg}    & \cmark & \cmark & \xmark \\
  & KG-R1 \citep{song2025efficient} & \xmark & \cmark & \cmark \\
  \colormidrule
  \rowcolor{blue!8}
  \multirow{1}{*}{\textbf{Ours}}
  & \textbf{GraphWalker}             & \cmark & \cmark & \cmark \\
  \colorbottomrule
  \end{tabular}
  \caption{Comparison of GraphWalker with existing agentic KGQA methods across three key properties: \textbf{\textit{Global KG}} (operating on global KGs rather than pre-extracted subgraphs), \textbf{\textit{Autonomous Interaction}} (interacting autonomously rather than via pre-defined workflows), and \textbf{\textit{RL Optimization}} (incorporating reinforcement learning to optimize reasoning policies).}
  \label{tab:method-comparison}
  \vspace{-14pt}
\end{table}

To address these issues, this paper proposes \textit{GraphWalker}, a framework that autonomously synthesizes interaction trajectories for \textit{Stage-Wise Fine-tuning}, followed by lightweight RL with a sparse exact-match (EM) reward.
Specifically, the two SFT stages directly target the aforementioned exploration and robustness bottlenecks. 
First, we conduct \textit{Constrained Random Walk (CRW)} on the KG and synthesize \textit{GraphSynth}, a corpus of 15k diverse trajectories that establishes a broad exploration prior.
Second, we construct \textit{GraphRoll}, a dataset of 6k distilled expert trajectories that equips the agent with reflection and error recovery capabilities.
Importantly, we demonstrate that the first stage serves as a
\textit{mid-training exploration scaffold}, which unlocks a higher performance ceiling during subsequent RL.

% To address these issues, this paper proposes \textit{GraphWalker}, a framework that autonomously synthesizes interaction trajectories for a \textit{Stage-Wise Fine-tuning}, followed by lightweight RL with a sparse exact-match reward.
% Specifically, the two SFT stages directly target the aforementioned exploration and robustness bottlenecks.
% First, to overcome the restricted search space caused by bypassing broad pre-training, a \textit{Constrained Random Walk (CRW)} synthesizes \textit{GraphSynth}---a corpus of 15k structurally diverse trajectories that establishes a broad exploration prior directly on the global KG.
% Second, to ensure robust navigation in massive, noisy environments, \textit{GraphRoll} provides 6k distilled expert trajectories that equip the agent with essential reflection and error recovery capabilities.
% Importantly, this paper demonstrates that the stage-wise SFT successfully builds the missing exploration foundation, unlocking a significantly higher performance ceiling for the final RL stage.

Our main contributions are summarized as follows:

\begin{itemize}[leftmargin=*]
    \vspace{-6pt}
    \item We identify the lack of an exploration prior as the core bottleneck in agentic KGQA generalization, and propose GraphWalker to address this limitation through two complementary datasets: \textit{GraphSynth}, 15k autonomously constructed, structurally diverse interaction trajectories, and \textit{GraphRoll}, 6k curated high-quality expert trajectories.
    
    \item We introduce a \textit{Stage-Wise SFT curriculum} to build a robust exploration foundation. By sequentially establishing a broad search prior and instilling error recovery, our method enables the agent to navigate vast, noisy KGs, unlocking a higher performance ceiling for subsequent RL optimization.
    
    \item Extensive experiments demonstrate that GraphWalker achieves SOTA performance on WebQSP and CWQ. Strong zero-shot results on GrailQA and our GraphWalkerBench confirm its superior robustness and generalization to unseen reasoning structures.
\end{itemize}

\section{GraphWalker Framework}

\begin{figure*}[t]
  \centering
  \includegraphics[width=\textwidth]{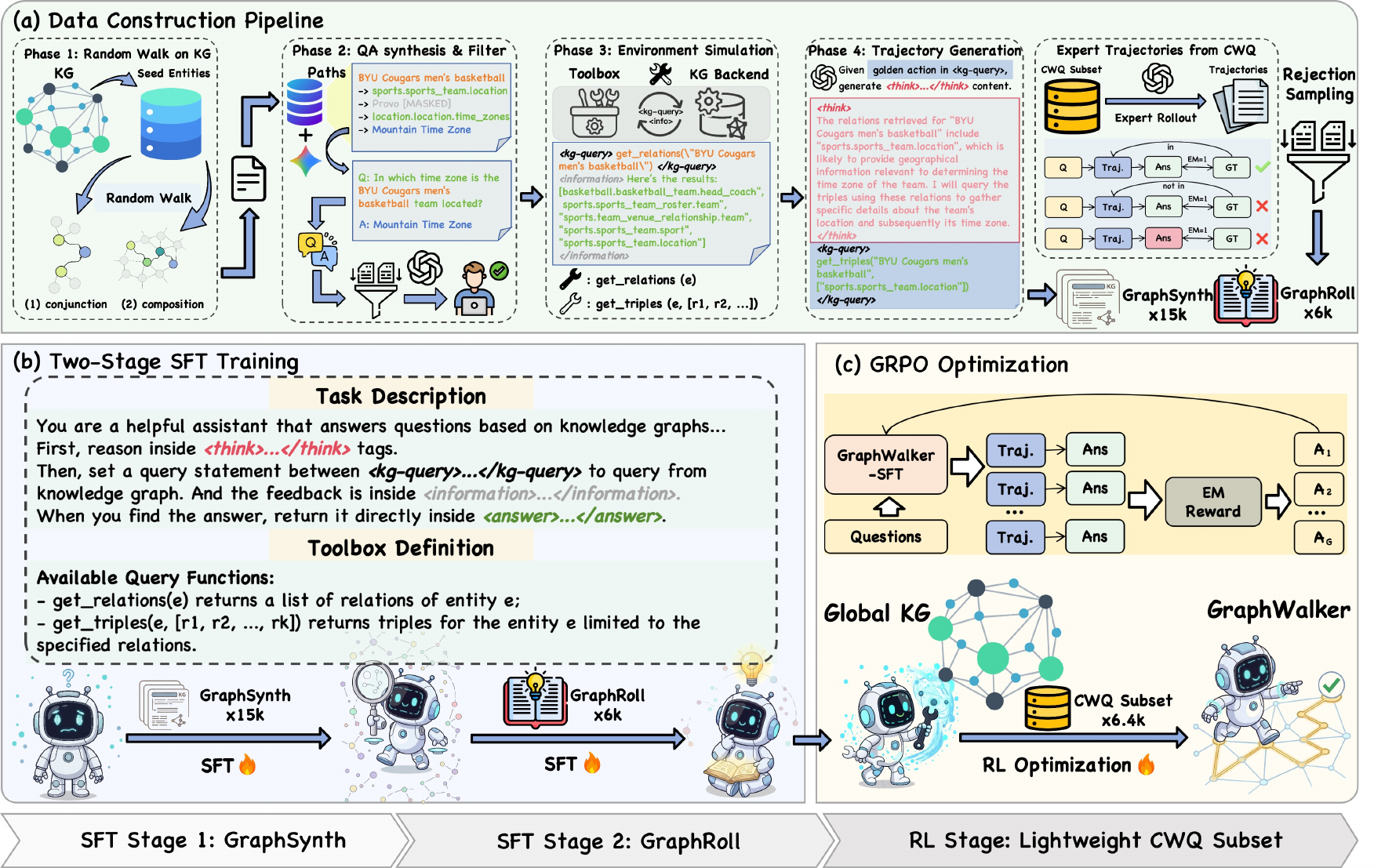}
  \vspace{-16pt}
  \caption{
  \textbf{Overview of our proposed GraphWalker.} 
  \textbf{(a) Data Construction:} a multi-phase pipeline for GraphSynth-15k and outcome-based rejection sampling for GraphRoll-6k. 
  \textbf{(b) Two-Stage SFT:} Stage 1 establishes a broad exploration prior, while Stage 2 instills reflection and error recovery. 
  \textbf{(c) RL Optimization:} GRPO with a simple exact-match reward unlocks a higher performance ceiling enabled by the two-stage SFT prior.}
  \label{fig:overview}
  \vspace{-14pt}
\end{figure*}

In this section, we present the GraphWalker framework, which consists of three components: the agentic environment, the data construction pipeline, and the training procedure. Figure~\ref{fig:overview} provides an overview of the full data construction and training pipeline.

\subsection{Agentic Initialization}
\label{sec:env_init}

\textbf{Environment.}
Following prior agentic KGQA systems \citep{sun2023think,jiang2025kg}, 
we formalize the KG as an executable environment backed by a \textit{SPARQL} endpoint. Unlike approaches that operate on pre-defined subgraphs, the agent interacts directly with the global KG in real time during both training and inference. 
This design exposes the agent to the full structural complexity of real-world KGs, including large branching factors and long-tail relations, which encourages more robust reasoning strategies.

\textbf{Toolbox.}
To interact with this environment, we equip the agent with two primitive tools:
(1)~$\textbf{get\_relations}(e)$ returns all relations incident 
to entity $e$, allowing the agent to survey the local neighborhood 
before selecting a retrieval direction.
(2)~$\textbf{get\_triples}(e, \mathcal{R}')$ returns all triples 
whose head or tail is $e$ and whose relation belongs to a subset 
$\mathcal{R}' \subseteq \mathcal{R}_{\{e\}}$ selected from the 
preceding \textbf{get\_relations} observation, enabling targeted 
and efficient KG traversal.

\textbf{Agent Loop.}
At each turn $t \leq T$, where $T$ is the maximum number of 
interaction turns, the agent is conditioned on the current state 
$s_t = (q, \tau_{t-1})$, where $q$ is the input question and 
$\tau_{t-1} = \{(c_i, a_i, o_i)\}_{i=1}^{t-1}$ is the interaction 
history up to the preceding turn. Here, $c_i$ denotes the model's 
internal reasoning process at turn $i$, $a_i$ the executed action, 
and $o_i$ the resulting observation.
Given $s_t$, the agent produces a response consisting of two parts: 
(1)~an internal reasoning process $c_t$, enclosed in 
\texttt{<think>}\ldots\texttt{</think>}, which reflects on the current 
state and determines the next action; and (2)~an action $a_t$, which 
is either a KG retrieval query enclosed in 
\texttt{<kg-query>}\ldots\texttt{</kg-query>}, or a final answer 
enclosed in \texttt{<answer>}\ldots\texttt{</answer>} when sufficient 
evidence has been gathered.
The executor processes $a_t$ on the KG $\mathcal{G}$ and returns the 
retrieval result as an observation $o_t$, enclosed in 
\texttt{<information>}\ldots\texttt{</information>} and appended to 
the trajectory. The full turn record $(c_t, a_t, o_t)$ is then 
appended to $\tau_t$.

\subsection{Data Construction}
\label{sec:data_construction}

In this section, we describe the construction of two complementary training corpora, \textbf{GraphSynth-15k} and \textbf{GraphRoll-6k}.
GraphSynth-15k focuses on broad structural coverage for exploration, while GraphRoll-6k emphasizes reflection and error recovery.

\subsubsection{GraphSynth: Diverse Trajectory Synthesis}

We first construct GraphSynth, a 15k corpus of KG-grounded interaction trajectories, using a four-phase pipeline designed to promote structural diversity, incorporate weak supervision, and simulate realistic interaction.

\textbf{Phase 1: Constrained Random Walk on KG.}

\begin{figure}[t]
  \centering
  \includegraphics[width=\linewidth]{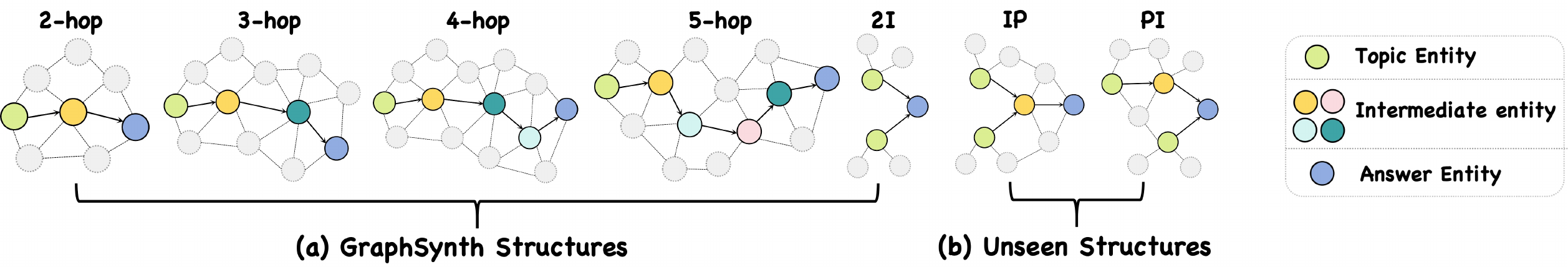}
  \caption{Representative interaction paths generated by constrained random walks.}
  \label{fig:crw_paths}
  \vspace{-14pt}
\end{figure}

We first extract the predicate set $\mathcal{P}$ and the seed entity set $\mathcal{E}$ from the SPARQL logical forms of the CWQ training split.
Despite using seen atomic relations, our CRW systematically recombines them into unobserved topological structures to drive compositional generalization.
Starting from a seed entity $e_0 \in \mathcal{E}$, we perform constrained random walks over relations in $\mathcal{P}$ to generate structurally diverse reasoning paths.
As shown in Figure~\ref{fig:crw_paths}(a), we introduce two complementary families of reasoning structures. \textbf{Composition chains} span \textbf{2--5 hops}, capturing sequential multi-hop exploration patterns. \textbf{Conjunction graphs} (denoted as \textbf{2I} in Figure~\ref{fig:crw_paths}(a)) originate from two distinct topic entities and converge on a shared target entity. 
Formally, each transition samples a relation--node pair subject to the predicate constraint:
\begin{equation}
r_t \in \mathcal{P}, \qquad e_{t+1}\in\mathcal{N}(e_t,r_t), \qquad d_{\min} \leq |\mathcal{N}(e_t,r_t)| \leq d_{\max},
\label{eq:crw_transition}
\end{equation}
where $\mathcal{N}(e_t,r_t)$ denotes the set of neighbors reachable from $e_t$ via relation $r_t$, and the cardinality constraint $d_{\min} \leq |\mathcal{N}(e_t,r_t)| \leq d_{\max}$ ensures that sampled relations are neither trivially unique nor excessively branching, thereby promoting structural diversity while avoiding degenerate or uninformative paths.

\textbf{Phase 2: Question Synthesis and Semantic Filtering.}

As shown in Figure~\ref{fig:overview} (Phase 2), for each reasoning path, we use \texttt{Gemini-2.5-Pro} \citep{comanici2025gemini25pushingfrontier} to generate a natural language question. 
To prevent information leakage, we \textbf{mask intermediate entities} (as $entity_1, entity_2, \dots$) along reasoning paths. For example, reasoning path "$\text{BYU Cougars men's basketball} \xrightarrow{\texttt{team.location}} entity_1 \xrightarrow{\texttt{time\_zones}} \text{Mountain Time Zone}$" yields: "In which time zone is the BYU Cougars men's basketball team located?"
To ensure data quality, we apply \textbf{LLM-based semantic filtering} using \texttt{GPT-4o-mini} \citep{hurst2024gpt}, which scores each question along three dimensions: 
(1) \textbf{reasoning path completeness}, 
(2) \textbf{question-path relevance}, 
and (3) \textbf{question semantic coherence}. 
Questions with an average score below $9$ out of $10$ are discarded.
In case of test-set leakage, we also verify that GraphSynth contains negligible semantic overlap with all the benchmarks (details in Appendix~\ref{app:contamination}).

To further validate the model's generalization capabilities, we also construct our own evaluation set, \textit{GraphWalkerBench},  which contains unseen reasoning structures in the GraphSynth dataset (shown in Figure~\ref{fig:crw_paths}(b), details in Appendix~\ref{app:datasets}).

\textbf{Phase 3: Environment Feedback Simulation.}

To faithfully simulate inference-time interaction, we execute each CRW path as a sequence of SPARQL queries against the full KG. We apply the same retrieval and filtering pipeline used at inference time, including \textbf{BM25}~\citep{bm25} candidate retrieval and \textbf{LLM-based relation reranking} (e.g., \texttt{Qwen2.5-7B-Instruct}). 
At each step $t$, we prompt an LLM to rerank the top-$k$ relations and up to 5 triples per relation as the observation $o_t$.

\textbf{Phase 4: Agent Thought Generation.}

As shown in Figure~\ref{fig:overview} (Phase 4), at each step $t$, given the current state $s_t$ and the corresponding oracle next action $a_{t}$, we use \texttt{GPT-4o-mini} as a rationale generator to generate the reasoning process $c_t$, i.e., chain-of-thought rationales that explain the selection of subsequent actions under the current context.
The resulting tuples $(c_t, a_t, o_t)$ are then assembled into complete interaction trajectories $\tau$, forming \textbf{GraphSynth-15k}.

\subsubsection{GraphRoll: Expert Trajectory Collection}
We then construct a smaller set of expert trajectories that demonstrate reflection and error recovery.
Given the complexity and diversity of CWQ, we sample a representative subset of 6k questions via stratified sampling across different question types to ensure balanced training coverage.
For each question, we use \texttt{GPT-4o-mini} as the expert model 
to generate high-quality trajectories that involve reflecting on 
uninformative observations and backtracking from dead ends.
We apply \textbf{outcome-based rejection sampling} based on two 
complementary criteria ensuring factual correctness and KG grounding:
\textbf{(1) Answer Correctness:} the predicted answer must be correct ($EM=1$).
\textbf{(2) Evidence Grounding:} all predicted answers must appear in the 
observation history $\{o_0, \ldots, o_T\}$, ensuring every answer is 
grounded in the retrieved KG evidence rather than parametric knowledge.
The resulting trajectories form \textbf{GraphRoll-6k}.

\subsection{GraphWalker Training}
\label{sec:training}

We adopt a two-stage SFT cold start followed by RL training. 
The first stage equips the agent with general exploration skills, 
the second cultivates reflection and error recovery, and RL 
subsequently aligns long-horizon decision making.
We parameterize the agent policy as an autoregressive language 
model $\pi_\theta$.
Specifically, in stage 1 and stage 2, the model is optimized using a next-token prediction objective over agentic interaction sequences, while treating \texttt{<information>} blocks as environmental context rather than learning targets:
\begin{equation}
\mathcal{L}(\theta) = 
-\sum_{t \in \mathcal{I}_{\text{agent}}} 
\log P_\theta(x_t \mid x_{<t}),
\end{equation}
where $\mathcal{I}_{\text{agent}}$ indexes tokens in the 
\texttt{<think>}, \texttt{<kg-query>}, and \texttt{<answer>} blocks.

\textbf{SFT Stage 1: Learning to Reason and Explore on Diverse KGs.}

In stage 1, we first train the agent on \textbf{GraphSynth-15k}. This stage aims to internalize robust navigation behaviors and 
reliable tool-invocation patterns under realistic retrieval noise, 
while treating KG retrieval results as environmental observations.

\newpage
\textbf{SFT Stage 2: Learning to Reflect and Recover from Errors.}

In stage 2, we fine-tune the model on \textbf{GraphRoll-6k}. 
Unlike stage 1, which emphasizes broad exploration, the expert trajectories in GraphRoll-6k contain explicit instances of error detection and reasoning recovery, guiding the model to develop more deliberate and self-corrective interaction strategies.

\textbf{RL Stage: Unlocking the Performance Ceiling}

Building on the two-stage SFT cold-start model, we apply 
RL to push the agent's performance beyond the SFT ceiling 
through autonomous exploration.
We train the model on a uniformly sampled CWQ subset using Group Relative Policy Optimization (GRPO) \citep{Guo_2025} with a sparse, trajectory-level exact match reward:
\begin{equation}
R(\tau) = s_{\text{EM}}(\tau) \in \{0, 1\}.
\end{equation}
Since our two-stage SFT already establishes an exploration and reflection prior, this sparse reward is sufficient to drive further policy improvement without dense intermediate supervision. The GRPO objective function is:

{\small
\begin{equation}
\mathcal{J}_{\text{GRPO}}(\theta) = \mathbb{E}_{q,\{\tau^{(i)}\}} \left[
  \frac{1}{N} \sum_{i=1}^{N} \frac{1}{|\tau^{(i)}|}
  \sum_{j=1}^{|\tau^{(i)}|}
  \min\!\left( \rho_{i,j} \hat{A}_{i,j},\;
  \mathrm{clip}(\rho_{i,j}, 1-\epsilon , 1+\epsilon)\, \hat{A}_{i,j} \right)
\right] - \beta\,\mathbb{D}_{\mathrm{KL}}(\pi_\theta \| \pi_{\text{ref}}),
\label{eq:grpo}
\end{equation}
}

where $\rho_{i,j}=\frac{\pi_\theta(a_{i,j}|s_{i,j})}{\pi_{\text{old}}(a_{i,j}|s_{i,j})}$ 
is the importance ratio, $\hat{A}_{i,j}$ is the group-normalized 
advantage, and $\pi_{\text{ref}}$ is the two-stage SFT model used as the KL reference policy.

\section{Experiments}

In this section, we conduct comprehensive experiments and report the experimental setup, main results, and ablation analyses. 
We answer the following research questions (RQs):

\textbf{RQ1:} Does GraphWalker outperform other LLM-based KGQA methods?
\textbf{RQ2:} Does GraphSynth broaden GraphWalker's search space and improve generalization?
\textbf{RQ3:} How does each training stage contribute to GraphWalker's final performance?
\textbf{RQ4:} How do the main components of our data construction pipeline work?

% ###################################################

\subsection{Experimental Setup}

\textbf{Datasets and Metrics.}
Following previous research \citep{sun2023think,luo2023reasoning}, we evaluate GraphWalker on two widely used KGQA benchmarks: CWQ \citep{talmor2018webknowledgebaseansweringcomplex} and WebQSP \citep{Yih2016TheVO}. To ensure a fair comparison, we utilize the gold topic entities provided by the benchmarks and report standard Exact Match (EM) and $F1$ scores \citep{talmor2018webknowledgebaseansweringcomplex}. Details of the datasets are described in Appendix~\ref{app:datasets}.

\textbf{Models and Baselines.}
We compare GraphWalker with seven previous prompting-based and training-based KGQA methods. 
% To ensure a fair comparison and reproducibility, we report the scores for these methods directly from their original publications.
To better evaluate the effectiveness of various methods on knowledge reasoning tasks, we use the full Freebase as a KG backend (\textit{Global KG} setting).
This setting is more challenging than approaches based on predefined subgraphs, as the agent must navigate the full KG.
For KG-R1, we evaluate the released checkpoint in the official repository under our Global KG setting. 
We additionally evaluate strong proprietary LLMs, including \texttt{DeepSeek-V3.2} and \texttt{GPT-4o-mini}, under the GraphWalker agent framework.
Details for all baselines are provided in Appendix~\ref{app:baselines}.

\textbf{Implementation Details.}
We train and evaluate GraphWalker on three open-source backbone models: \textbf{Qwen2.5-3B-Instruct, Qwen2.5-7B-Instruct} \citep{qwen2025qwen25technicalreport}, and \textbf{Llama-3.1-8B-Instruct} \citep{dubey2024llama}.  
For SFT data collection, we use \texttt{GPT-4o-mini} to generate trajectories for both GraphSynth and GraphRoll. For evaluation, we set the max turn number $T = 10$.
Our training implementation details are shown in Appendix~\ref{app:training_details}.

% ###################################################

\subsection{Main Results (RQ1)}

\begin{table*}[t]
  \centering
  \scriptsize
  \setlength{\tabcolsep}{10pt}
  \renewcommand{\arraystretch}{1.1}
  \begin{tabular}{l|l|cc|cc}
  \toprule
  \multirow{2}{*}{\textbf{Method}}
  & \multirow{2}{*}{\textbf{Backbone}}
  & \multicolumn{2}{c|}{\textbf{CWQ}}
  & \multicolumn{2}{c}{\textbf{WebQSP}} \\
  & & \textbf{EM} & \textbf{F1} & \textbf{EM} & \textbf{F1} \\
  \colormidrule
  \rowcolor{gray!15}
  \multicolumn{6}{c}{\textit{Vanilla LLMs}} \\
  \colormidrule
  \rowcolor{gray!3}
  IO Prompt & Qwen2.5-3B-Instruct & 22.0 & 17.7 & 44.6 & 30.3 \\
  \rowcolor{gray!3}
  IO Prompt & Qwen2.5-7B-Instruct & 25.7 & 20.7 & 50.9 & 33.2 \\
  \rowcolor{gray!3}
  IO Prompt & GPT-4o-mini & 45.5 & 33.6 & 47.1 & 39.3 \\
  \rowcolor{gray!3}
  IO Prompt & DeepSeek-V3.2 & 50.1 & 43.5 & 63.8 & 55.7 \\
  \colormidrule
  \rowcolor{gray!15}
  \multicolumn{6}{c}{\textit{Agentic KGQA Methods}} \\
  \colormidrule
  \rowcolor{gray!7}
  RoG & LLaMA-2-7B-Instruct & 62.6 & 56.2 & 85.7 & 70.8 \\
  \rowcolor{gray!7}
  ToG & GPT-4 & 69.5 & - & 81.9 & - \\
  \rowcolor{gray!7}
  ToG-2.0 & GPT-3.5 & 68.9 & 65.8 & 77.8 & 74.5 \\
  \rowcolor{gray!7}
  GoG & GPT-4 & 75.2 & - & 84.4 & - \\
  \rowcolor{gray!7}
  KBQA-o1 & LLaMA3.1-8B-Instruct& - & - & 75.8 & 82.1 \\
  \rowcolor{gray!7}
  KG-Agent & LLaMA2-7B-Instruct & 72.2 & \underline{69.8} & 83.3 & 81.0 \\
  \rowcolor{gray!7}
  ${}^{\dagger}$KG-R1 & Qwen2.5-3B-Instruct & 66.8 & 61.7 & 82.1 & 78.9 \\
  \colormidrule
  \rowcolor{gray!15}
  \multicolumn{6}{c}{\textit{GraphWalker (Our Method)}} \\
  \colormidrule
  \rowcolor{gray!11}
  ${}^{\dagger}$Vanilla Agent
    & Qwen2.5-7B-Instruct
    & 40.7 & 33.2
    & 68.4 & 66.1 \\
  \rowcolor{gray!11}
  ${}^{\dagger}$Vanilla Agent
    & GPT-4o-mini
    & 63.4 & 60.3
    & 79.6 & 70.6 \\
  \rowcolor{gray!11}
  ${}^{\dagger}$Vanilla Agent
    & DeepSeek-V3.2
    & 69.8 & 63.5
    & 76.7 & 71.8 \\
  \rowcolor{blue!8}
  GraphWalker-7B-SFT
    & Qwen2.5-7B-Instruct
    & 68.3 & 63.2 & 82.0 & 79.1 \\
  \rowcolor{blue!8}
  GraphWalker-3B-SFT-RL
    & Qwen2.5-3B-Instruct
    & 70.9 & 65.2 & 83.5 & 81.7 \\
  \rowcolor{blue!8}
  GraphWalker-8B-SFT-RL
    & LLaMA3.1-8B-Instruct
    & \underline{78.5} & 69.6 & \underline{88.2} & \underline{84.5} \\
  \rowcolor{blue!15}
  \textbf{GraphWalker-7B-SFT-RL}
    & Qwen2.5-7B-Instruct
    & \textbf{79.6} & \textbf{74.2}
    & \textbf{91.5} & \textbf{88.6} \\
  \colorbottomrule
  \end{tabular}
  \vspace{-6pt}
  \caption{Performance comparison (in percentage) of KGQA methods on CWQ and WebQSP. The best and second-best results are shown in \textbf{bold} and \underline{underlined}, respectively.~ ${}^{\dagger}$ denotes models evaluated under the GraphWalker interaction framework with full global KG.}
  \label{tab:main-results}
  \vspace{-16pt}
\end{table*}

As shown in Table~\ref{tab:main-results}, GraphWalker achieves SOTA performance across all benchmarks, demonstrating its effectiveness from two perspectives:

\textbf{Superiority in Complex Reasoning.} 
GraphWalker consistently outperforms both prompting and training baselines.
Notably, it surpasses the GPT-4-powered GoG by 4.4\% EM on CWQ, and outperforms training-based models such as KG-Agent (+7.4\% EM on CWQ) and KBQA-o1 (+15.7\% EM on WebQSP).
Under the more challenging Global KG setting, GraphWalker-3B-SFT-RL outperforms KG-R1 (same \texttt{Qwen2.5-3B-Instruct} backbone) by a clear margin (e.g., +4.1\% EM on CWQ), suggesting that KG-R1's reliance on pre-extracted subgraphs undermines its robustness.
Moreover, equipping proprietary LLMs (\texttt{GPT-4o-mini} and \texttt{DeepSeek-V3.2}) with our interaction framework yields substantial gains over their IO Prompt counterparts (+17.9\% and +19.7\% EM on CWQ). 
Impressively, GraphWalker with trained backbones outperforms these proprietary LLMs across all model sizes, demonstrating the effectiveness of our synthetic agentic curriculum and training strategy.

\textbf{Cross-task Transferability.} Despite being trained solely on CWQ, GraphWalker effectively cross-task transfers to WebQSP (91.5\% EM), outperforming all baselines trained directly on WebQSP data. This confirms that training on CWQ's structurally diverse, high-complexity paths equips the agent with highly transferable exploration strategies.

% ###################################################
\subsection{Comparison Analysis (RQ2)}

\begin{wrapfigure}{r}{0.36\linewidth}
    \centering
    \vspace{-3.6em}
    \includegraphics[width=\linewidth]{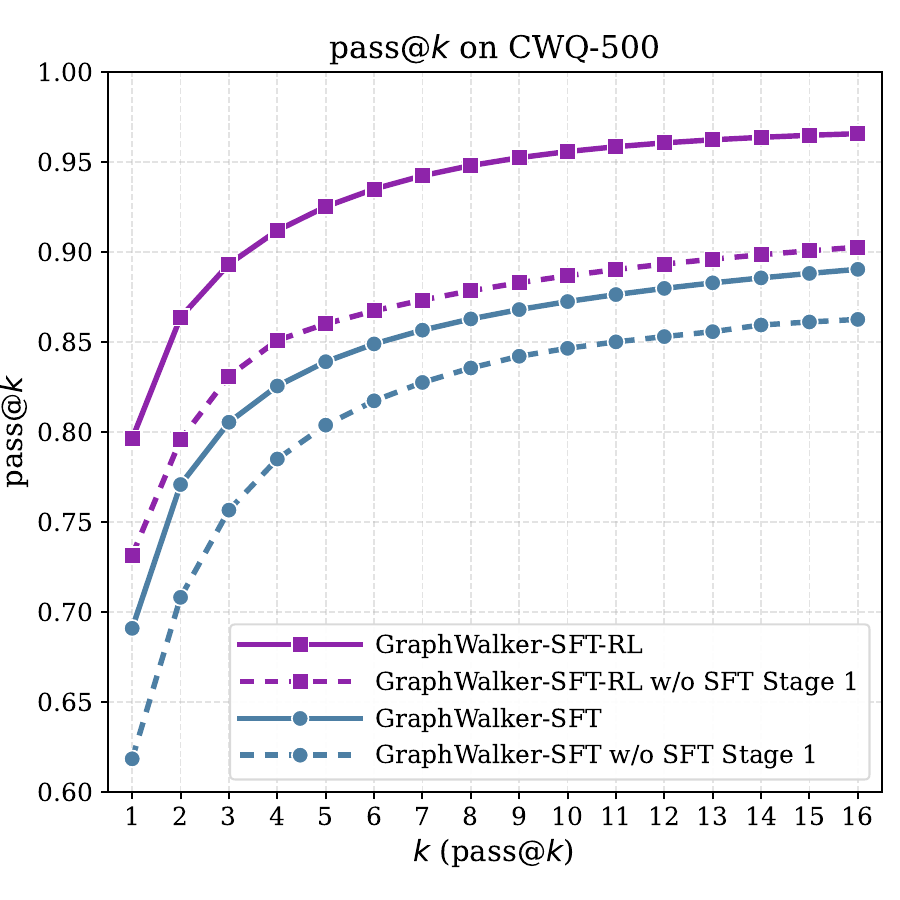}
    \vspace{-2.0em}
    \caption{Pass@$k$ comparison.}
    \label{fig:passk}
    \vspace{-1.6em}
\end{wrapfigure}

\textbf{Pass@$k$ Analysis.}
To validate that GraphSynth effectively broadens the agent's reasoning search space, we sample 500 questions from the CWQ test set and evaluate exploration capacity via \textbf{Pass@$k$} over $32$ runs.
As shown in Figure~\ref{fig:passk}, within both the SFT-only and SFT-RL settings, removing GraphSynth (\textbf{w/o SFT Stage 1}) consistently reduces Pass@$k$ across all $k$, confirming its critical role in expanding the agent's exploration capacity. In contrast, models trained with GraphSynth exhibit higher Pass@$k$ values as $k$ increases, indicating that our CRW synthesis establishes a broader exploration prior. 
Furthermore, \textbf{GraphWalker-SFT-RL} achieves the highest Pass@$k$ across all $k$, suggesting that a stronger exploration foundation enables RL to further amplify these gains.

\textbf{Comparison with GPT-4o-mini Best-of-N.}
We further evaluate \texttt{GPT-4o-mini} with Best-of-N sampling in the same GraphWalker scaffold on the full CWQ test set. With an oracle selector, its Best-of-8 reaches 81.70\% EM, whereas GraphWalker reaches 93.73\%. GraphWalker also consistently achieves higher Pass@$k$ at $k=2,4,8,16$ (86.38\%, 91.19\%, 94.80\%, and 96.57\%) than \texttt{GPT-4o-mini} (71.91\%, 77.85\%, 82.78\%, and 86.97\%). This result shows that the teacher has non-trivial trajectory diversity, while GraphWalker produces a substantially broader set of successful trajectories. Unlike oracle Best-of-N selection, our RL stage directly increases the likelihood of successful trajectories during standard inference.

\textbf{Zero-shot Generalization.}
To assess whether GraphSynth improves reasoning generalization, we evaluate GraphWalker and its ablated variants on two zero-shot benchmarks: the zero-shot split of \textit{GrailQA} and \textit{GraphWalkerBench}, a structurally diverse evaluation set involving unseen reasoning structures not present in GraphSynth (see Appendix~\ref{app:datasets}).
As shown in Table~\ref{tab:zeroshot}, both SFT stages are crucial for out-of-distribution (OOD) generalization. Removing GraphSynth (\textbf{w/o SFT Stage 1}) significantly impairs generalization (6.80\% average EM drop), confirming its essential role in broadening the agent's search space. 
Similarly, removing GraphRoll (\textbf{w/o SFT Stage 2}) leads to substantial performance degradation (11.10\% average EM drop), as the model's ability to self-correct is significantly weakened. Furthermore, the consistent improvement of \textit{GraphWalker-Full} over \textit{w/o RL} confirms that RL further enhances the generalization capability established by the SFT stages.

\begin{table}[t]
\centering
\scriptsize
\setlength{\tabcolsep}{3pt}
\renewcommand{\arraystretch}{1.1}
\begin{minipage}[t]{0.48\linewidth}
\centering
\begin{tabular}{cccccc}
\toprule
\multirow{2}{*}{Method} & \multicolumn{2}{c}{GrailQA} 
& \multicolumn{2}{c}{\makecell{GraphWalker\\Bench}}
& \multirow{2}{*}{$\Delta~\overline{\text{EM}}$} \\
\cmidrule(lr){2-3} \cmidrule(lr){4-5}
& EM & F1 & EM & F1 & \\
\colormidrule
\rowcolor{blue!8}
\textbf{GraphWalker-Full} & \textbf{86.3} & \textbf{84.4} & \textbf{63.5} & \textbf{60.8} & --- \\
\colormidrule
\rowcolor{gray!15}
\multicolumn{6}{c}{\textit{Ablations}} \\
\colormidrule
w/o SFT Stage 1 & 80.9 & 78.0 & 55.3 & 52.4 & \textcolor{green!60!black}{$-$6.80} \\
w/o SFT Stage 2 & 74.3 & 69.7 & 53.3 & 49.5 & \textcolor{green!60!black}{$-$11.10} \\
w/o RL          & 79.4 & 76.3 & 49.7 & 47.0 & \textcolor{green!60!black}{$-$10.35} \\
\bottomrule
\end{tabular}
\caption{Zero-shot performance on \textit{Grail-}\\\textit{QA} and \textit{GraphWalkerBench}, which involves unseen structures in \textit{GraphSynth}.}
\label{tab:zeroshot}
\end{minipage}
\hfill
\begin{minipage}[t]{0.50\linewidth}
\centering
\begin{tabular}{ccccccc}
\toprule
\multirow{2}{*}{Method} & \multicolumn{2}{c}{CWQ} & \multicolumn{2}{c}{WebQSP}
& \multirow{2}{*}{$\Delta~\overline{\text{EM}}$} \\
\cmidrule(lr){2-3} \cmidrule(lr){4-5}
& EM & F1 & EM & F1 & \\
\colormidrule
\rowcolor{blue!8}
\textbf{GraphWalker-Full}
& \textbf{79.6} & \textbf{74.2}
& \textbf{91.5} & \textbf{88.6} & --- \\
\colormidrule
\rowcolor{gray!15}
\multicolumn{6}{c}{\textit{Training Stage Ablations}} \\
\colormidrule
w/o SFT Stage 1 & 75.2 & 70.4 & 85.7 & 82.5 & \textcolor{green!60!black}{$-$5.10} \\
w/o SFT Stage 2 & 72.3 & 66.3 & 80.8 & 77.5 & \textcolor{green!60!black}{$-$9.00} \\
w/o SFT         & 70.6 & 66.0 & 79.8 & 75.6 & \textcolor{green!60!black}{$-$10.35} \\
w/o RL          & 68.3 & 63.2 & 82.0 & 79.1 & \textcolor{green!60!black}{$-$10.40} \\
\colormidrule
\rowcolor{gray!15}
\multicolumn{6}{c}{\textit{Training Order Ablation}} \\
\colormidrule
Mixed-SFT+RL  & 70.2 & 64.2 & 80.1 & 75.8 & \textcolor{green!60!black}{$-$10.40} \\
\bottomrule
\end{tabular}
\caption{Ablation study on \textit{CWQ} and \textit{WebQSP}.}
\label{tab:phase_ablation}
\end{minipage}
\vspace{-10pt}
\end{table}

% ###################################################

\subsection{Analysis on Each Training Stage (RQ3)}

\textbf{Ablation of Each Stage.}
To understand the contribution of each training stage, we ablate 
individual components from the training pipeline.
As shown in Table~\ref{tab:phase_ablation}, removing GraphSynth (\textbf{w/o SFT Stage 1}) causes a substantial decline (5.10\% average EM drop), indicating its critical role in establishing broad exploration priors for navigating complex KG structures. 
Meanwhile, removing GraphRoll (\textbf{w/o SFT Stage 2}) leads to an even greater drop (9.00\% average EM drop), demonstrating that 
GraphRoll's expert trajectories are essential for instilling reflection and self-correction capabilities. 
Finally, removing RL results in a severe 10.40\% average 
EM drop, confirming that RL further consolidates the capabilities 
established by the SFT stages.

\textbf{Importance of Stage-wise Training.}
To validate the proposed curriculum, we compare stage-wise training against a \textbf{Mixed} baseline combining GraphSynth and GraphRoll into a single SFT pass with identical RL training. As shown in Table~\ref{tab:phase_ablation}, mixed training consistently underperforms (dropping from 79.6\% to 70.2\% EM on CWQ) despite using identical data, confirming that training order is critical: 
Stage 1 must first establish broad exploration priors before Stage 2 
can effectively instill reflection and recovery capabilities.

\textbf{Behavioral Changes after RL.}
The RL stage is not designed to learn KG navigation from scratch with sparse rewards alone. Instead, the two-stage SFT model already makes rewardable exploration and recovery trajectories reachable, and RL shifts probability mass toward trajectories with more complete evidence acquisition and verification. To characterize this change, we compare the trajectories produced before and after RL across all evaluation sets.

\begin{figure*}[t]
\centering
\begin{subfigure}[b]{0.196\textwidth}
  \centering
  \includegraphics[width=\linewidth]{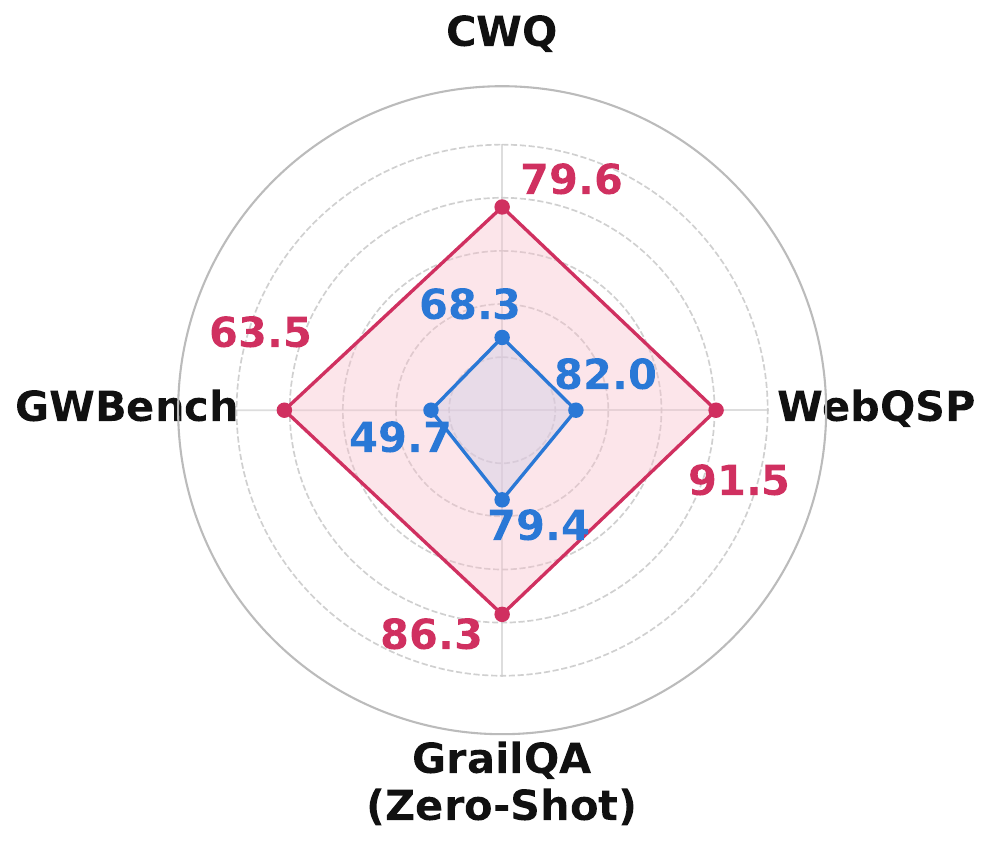}
  \caption{EM}
  \label{fig:radar_em}
\end{subfigure}\hspace{1pt}%
\begin{subfigure}[b]{0.196\textwidth}
  \centering
  \includegraphics[width=\linewidth]{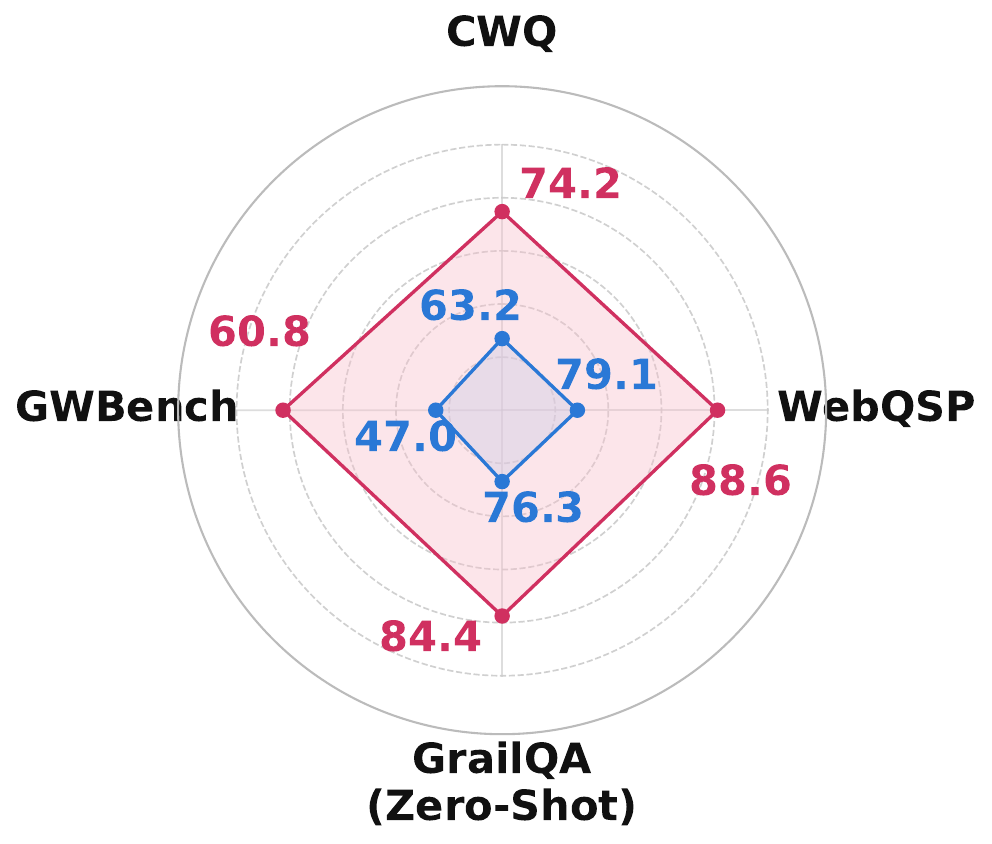}
  \caption{F1}
  \label{fig:radar_f1}
\end{subfigure}\hspace{1pt}%
\begin{subfigure}[b]{0.196\textwidth}
  \centering
  \includegraphics[width=\linewidth]{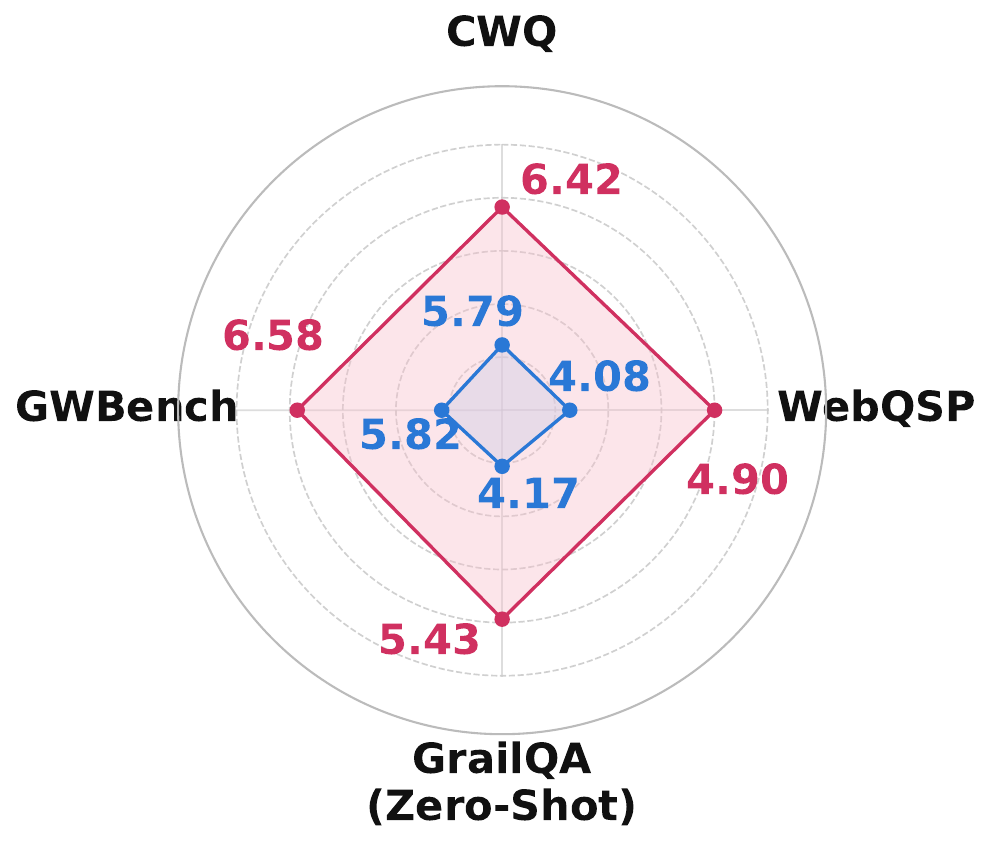}
  \caption{Avg.\ Turns}
  \label{fig:radar_turns}
\end{subfigure}\hspace{1pt}%
\begin{subfigure}[b]{0.196\textwidth}
  \centering
  \includegraphics[width=\linewidth]{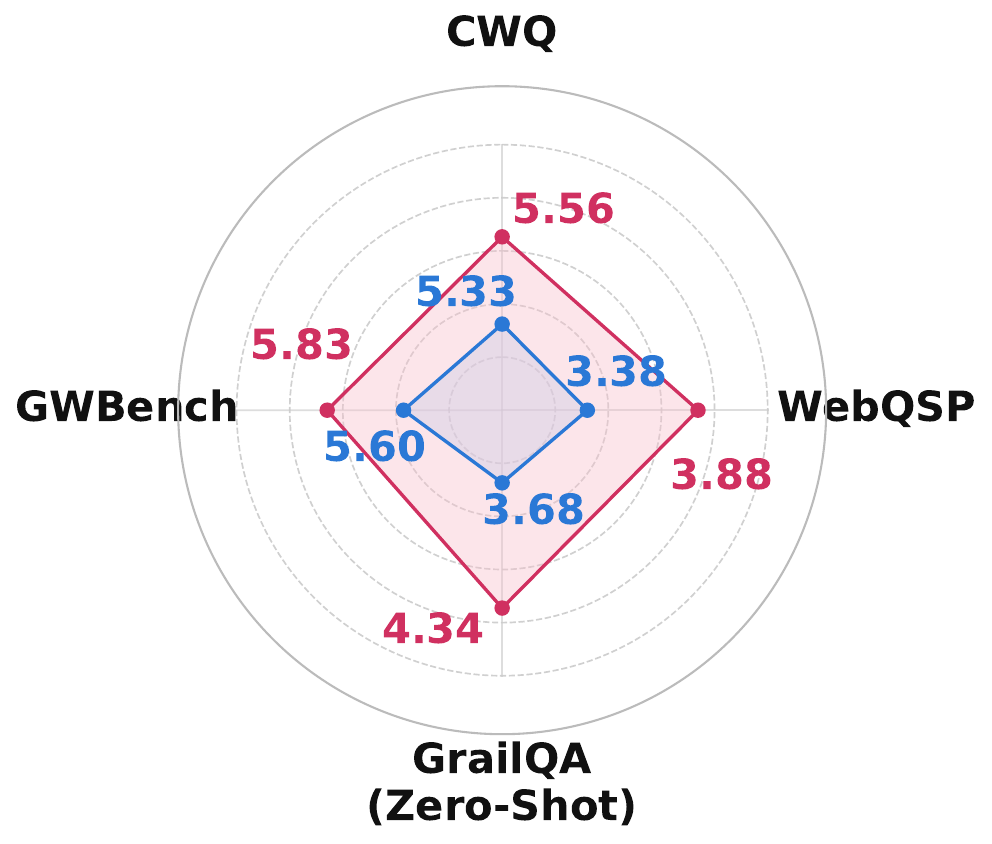}
  \caption{Avg.\ Tool-calls}
  \label{fig:radar_calls}
\end{subfigure}\hspace{1pt}%
\begin{subfigure}[b]{0.196\textwidth}
  \centering
  \includegraphics[width=\linewidth]{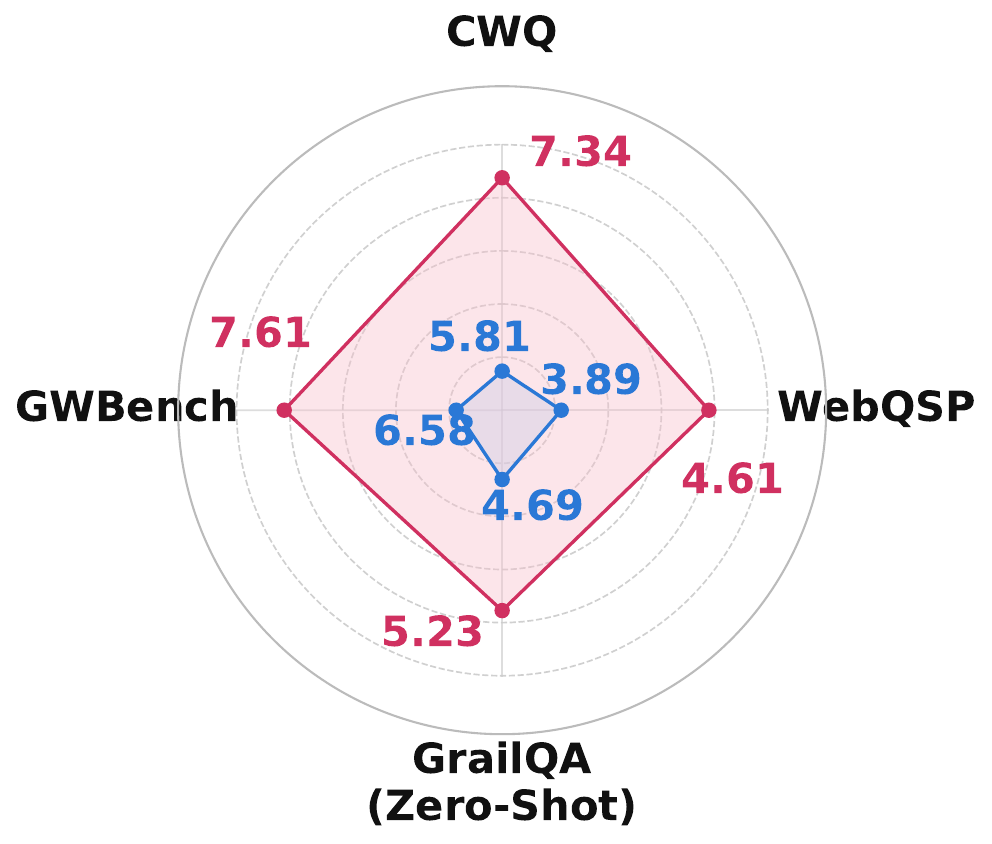}
  \caption{Avg.\ Reflections}
  \label{fig:radar_refl}
\end{subfigure}
\\[4pt]
{\small
  \raisebox{2pt}{\colorbox[RGB]{184,216,248}{\hspace{10pt}}}\hspace{3pt}GraphWalker w/o RL%
  \hspace{20pt}%
  \raisebox{2pt}{\colorbox[RGB]{248,192,204}{\hspace{10pt}}}\hspace{3pt}GraphWalker%
}
\vspace{-4pt}
\caption{Behavioral comparison between \textit{GraphWalker} and \textit{GraphWalker w/o RL} across all benchmarks. Each axis is independently scaled to highlight per-benchmark differences. Fine-grained per-reasoning-type EM analysis is shown in Appendix~\ref{app:rl_radar_bytype}.}
\label{fig:rl_behavior_radar}
\vspace{-10pt}
\end{figure*}

As shown in Figure~\ref{fig:rl_behavior_radar}, RL leads to higher EM and F1 scores across all benchmarks, with longer exploration trajectories (Avg.\ Turns) and more tool calls, while reflection and recovery behaviors also become more frequent. These changes suggest that the performance gain comes primarily from more complete evidence acquisition and verification rather than merely improved tool syntax or earlier stopping. Fine-grained per-type analysis (Appendix~\ref{app:rl_radar_bytype}) further shows that the largest gain occurs on IP (+32.05 EM), confirming that RL especially strengthens evidence acquisition on structurally challenging queries.

% ##############################################################

\subsection{Ablation Study (RQ4)}

To better understand the internal mechanisms of GraphWalker, we conduct ablation experiments on key components of our pipeline.

\begin{wrapfigure}{r}{0.48\linewidth}
    \centering
    \vspace{-1.2em}
    \includegraphics[width=\linewidth]{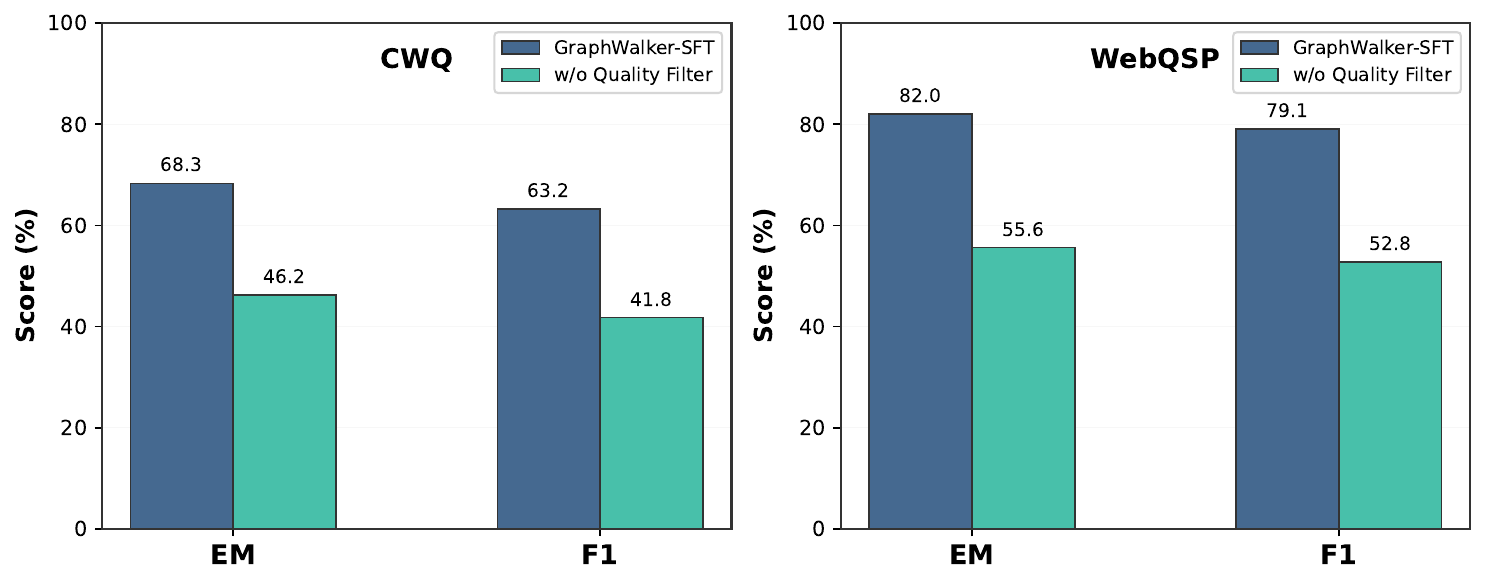}
    \vspace{-2.0em}
    \caption{Ablation study on the \textit{quality filter}.}
    \label{fig:ablation}
\end{wrapfigure}

\textbf{Impact of Quality Filtering.}
As shown in Figure~\ref{fig:ablation}, removing the \textit{quality filter} from the GraphWalker-SFT model causes substantial performance drops on both datasets: EM drops from 68.3\% to 46.2\% on CWQ and from 82.0\% to 55.6\% on WebQSP, confirming that filtering out low-quality questions is critical to the overall data pipeline.

% ###################################################

\textbf{Impact of Data Scale.}

\begin{figure}[ht]
    \centering
    \includegraphics[width=0.90\linewidth]{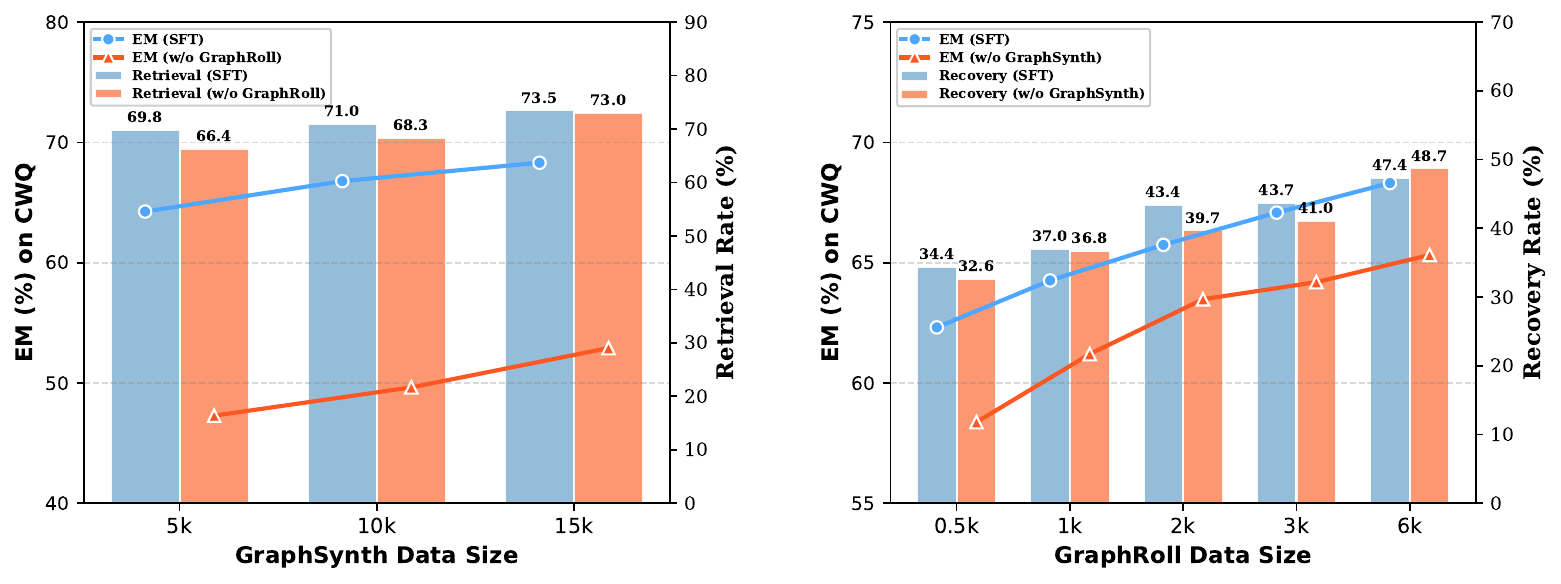}
    \caption{
    Impact of data scale on performance.
    \textit{Left}: scaling GraphSynth data improves \textit{EM} and \textit{Retrieval Rate}.
    \textit{Right}: scaling GraphRoll data improves \textit{EM} and \textit{Recovery Rate}.
    }
    \label{fig:scaling}
    \vspace{-10pt}
\end{figure}

We investigate the effect of data scale on each training stage separately.
For Stage 1, we vary the size of GraphSynth across $\{5\text{k}, 10\text{k}, 15\text{k}\}$ samples; for Stage 2, we vary GraphRoll across $\{0.5\text{k}, 1\text{k}, 2\text{k}, 3\text{k}, 6\text{k}\}$ samples. 
Here, \textit{Retrieval Rate} measures the proportion of test cases where the gold answer appears in the agent's retrieved observations, reflecting the agent's ability to find relevant evidence through KG exploration; 
\textit{Recovery Rate} measures the proportion of cases where the 
agent successfully corrects an erroneous \texttt{<kg-query>} step 
and recovers to the correct answer.
As shown in Figure~\ref{fig:scaling}, both \textit{EM} and the corresponding \textit{Retrieval/Recovery Rates} improve consistently with data scale, confirming the benefit of scaling trajectory supervision. 
Notably, the performance gap between the full model and the ablated variants (\textbf{w/o GraphRoll} and \textbf{w/o GraphSynth}) persists across all scales, demonstrating that the two stages provide complementary and irreplaceable contributions.

% ###################################################

\section{Related work}
\label{sec:related_work}

\textbf{Knowledge Graph Question Answering (KGQA).}
To address the limitations of unstructured text retrieval and LLM hallucination, KGs have been widely adopted in question answering tasks for their ability to provide structured world knowledge \citep{baek2023knowledgeaugmentedlanguagemodelverification, xu2025knowledgeinfusedpromptingassessingadvancing}.
Traditional KGQA methods fall into two categories: \textbf{Information retrieval (IR)} methods \citep{Zhang_2022, dong2023bridgingkbtextgapleveraging, li2024chainofknowledgegroundinglargelanguage} retrieve subgraphs from KGs and extract answers from the gathered evidence, while \textbf{Semantic parsing (SP)} methods \citep{berant2013semantic, yih2015semantic, li2023fewshotincontextlearningknowledge, Luo_2024} translate questions into logical forms (e.g., SPARQL) for verifiable execution.
However, both are brittle under noisy or incomplete KG schemas and struggle to handle long-tail entities, motivating the shift toward agentic paradigms.

\textbf{Agentic KGQA.}
Agentic KGQA treats LLMs as autonomous agents that interact with 
KGs through iterative exploration and retrieval.
\textbf{Prompting-based} methods such as ToG \citep{sun2023think} and GoG \citep{xu2024generate} perform iterative graph traversal but rely on predefined workflows or costly closed-source LLMs.
\textbf{Fine-tuning} methods learn from curated training trajectories: CoT-based approaches \citep{wu2025cotkrchainofthoughtenhancedknowledge} enhance step-by-step reasoning transparency, RoG \citep{luo2023reasoning} grounds plans in KGs, and KG-Agent \citep{jiang2025kg} employs multi-agent reasoning but 
remains limited to synthesized program data.
\textbf{Reinforcement learning} methods such as KG-R1 \citep{song2025efficient} optimize the retrieval policy end-to-end but still rely on subgraph pre-extraction.

\section{Conclusion}
In this paper, we propose GraphWalker, a novel agentic KGQA framework that addresses training data scarcity and generalization bottlenecks via automated trajectory synthesis and stage-wise fine-tuning. 
Through our two-stage SFT curriculum, the agent first establishes a broad exploration prior over diverse KG structures, and subsequently acquires critical reflection and error recovery capabilities. Notably, this robust foundation unlocks a higher performance ceiling for subsequent reinforcement learning (RL) optimization.
Extensive experiments demonstrate that GraphWalker achieves 
state-of-the-art performance and strong zero-shot generalization 
to unseen reasoning topologies, confirming its effectiveness and 
scalability for autonomous reasoning over global KGs.

% \section*{Author Contributions}

% \section*{Acknowledgments}
% Use unnumbered first level headings for the acknowledgments. All

% \section*{Ethics Statement}
% Authors can add an optional ethics statement to the paper. 
% For papers that touch on ethical issues, this section will be evaluated as part of the review process. The ethics statement should come at the end of the paper. It does not count toward the page limit, but should not be more than 1 page. 

\bibliography{colm2026_conference}
\bibliographystyle{colm2026_conference}

\newpage
\appendix

\section{Preliminaries}

\textbf{Knowledge graph.}
A knowledge graph (KG) $\mathcal{G} = (\mathcal{E}, \mathcal{R}, \mathcal{T})$ 
consists of a set of entities $\mathcal{E}$, a set of relations 
$\mathcal{R}$, and a triple set $\mathcal{T} \subseteq \mathcal{E} 
\times \mathcal{R} \times \mathcal{E}$, where each triple 
$\langle e, r, e' \rangle \in \mathcal{T}$ encodes a directed 
relation $r$ from entity $e$ to entity $e'$. Each entity is 
assigned a unique entity ID (e.g., \texttt{m.02mjmr} for 
\emph{Barack Obama}) and belongs to one or more entity types 
such as \emph{Country} or \emph{Person}.
Furthermore, we introduce \textit{neighboring relations} to denote both the incoming and outgoing relations for a set of entities $\{e\}$, denoted as $\mathcal{R}_{\{e\}} = \{r \mid \langle e, r, e' \rangle \in \mathcal{G}\} \cup \{r \mid \langle e', r, e \rangle \in \mathcal{G}\}$.

\textbf{Reasoning Paths.}
A reasoning path is an instance of a relation path $z$ in the 
KG: $w_z = e_0 \xrightarrow{r_1} e_1 \xrightarrow{r_2} \cdots 
\xrightarrow{r_l} e_l$, where $e_i \in \mathcal{E}$ denotes the 
$i$-th entity and $r_i$ denotes the $i$-th relation in $z$.

\textbf{Example.} Given a relation path 
$z = \texttt{marry\_to} \rightarrow \texttt{father\_of}$, 
a reasoning path instance could be:
$w_z = \text{Alice} \xrightarrow{\texttt{marry\_to}} 
\text{Bob} \xrightarrow{\texttt{father\_of}} \text{Charlie}$, which denotes ``Alice'' is married to ``Bob'' and ``Bob'' is the father of ``Charlie''.

\textbf{Problem Formulation.} In this work, we assume that a KG is available and contains the answer entities for the given natural language question. Our objective is to develop an LLM-based agent that can autonomously infer the answer to the question based on the given KG. As it has been shown that an interface is helpful for LLMs to manipulate structured data \citep{jiang2023structgpt}, we further assume that a toolbox can be provided to facilitate access to the information of the KG. Formally, given a natural language question $q$, topic entities $\mathcal{E}_q \subseteq \mathcal{E}$ obtained via entity linking, a toolbox $\mathcal{T}$, and a KG $\mathcal{G}$, we aim to develop a capable agent that interacts with the KG through iterative tool calls \citep{yao2023reactsynergizingreasoningacting, jiang2025kg, Xiong_2024} to deduce the final answers $\widehat{\mathcal{A}}_q \subseteq \mathcal{E}$ for the question $q$.

\section{Data Construction Details}
\label{app:data_construction}
\subsection{Constrained Random Walk Algorithm}
\label{crw_alg}
\begin{algorithm}[ht]
\caption{Constrained Random Walk (CRW)}
\label{alg:crw}
\begin{algorithmic}[1]
\STATE \textbf{Input:} KG executor, seed set $\mathcal{E}$, predicate set $\mathcal{P}$, target size $N$
\STATE \textbf{Output:} trajectory set $\mathcal{D}_{\text{crw}}$
\STATE $\mathcal{D}_{\text{crw}}\leftarrow\emptyset$
\WHILE{$|\mathcal{D}_{\text{crw}}| < N$}
  \STATE Sample seed entity $e_0 \sim \mathcal{E}$ and hop length $L$
  \STATE Initialize partial path $\pi\leftarrow[e_0]$
  \FOR{$t=0$ to $L-1$}
    \STATE Retrieve outgoing candidate relations of $e_t$, then keep $\mathcal{R}_t=\mathcal{R}(e_t)\cap\mathcal{P}$
    \IF{$\mathcal{R}_t=\emptyset$}
      \STATE \textbf{break}
    \ENDIF
    \STATE Sample relation $r_t\sim \mathcal{R}_t$, then sample next node $e_{t+1}\sim \mathcal{N}(e_t,r_t)$
    \STATE Append $(r_t,e_{t+1})$ to $\pi$
  \ENDFOR
  \IF{$\pi$ is valid}
    \STATE Add $\pi$ to $\mathcal{D}_{\text{crw}}$
  \ENDIF
\ENDWHILE
\STATE \textbf{return} $\mathcal{D}_{\text{crw}}$
\end{algorithmic}
\end{algorithm}

\subsection{Semantic Contamination Analysis}
\label{app:contamination}

\begin{figure*}[t]
    \centering
    \begin{subfigure}[b]{0.22\linewidth}
        \centering
        \includegraphics[width=\linewidth]{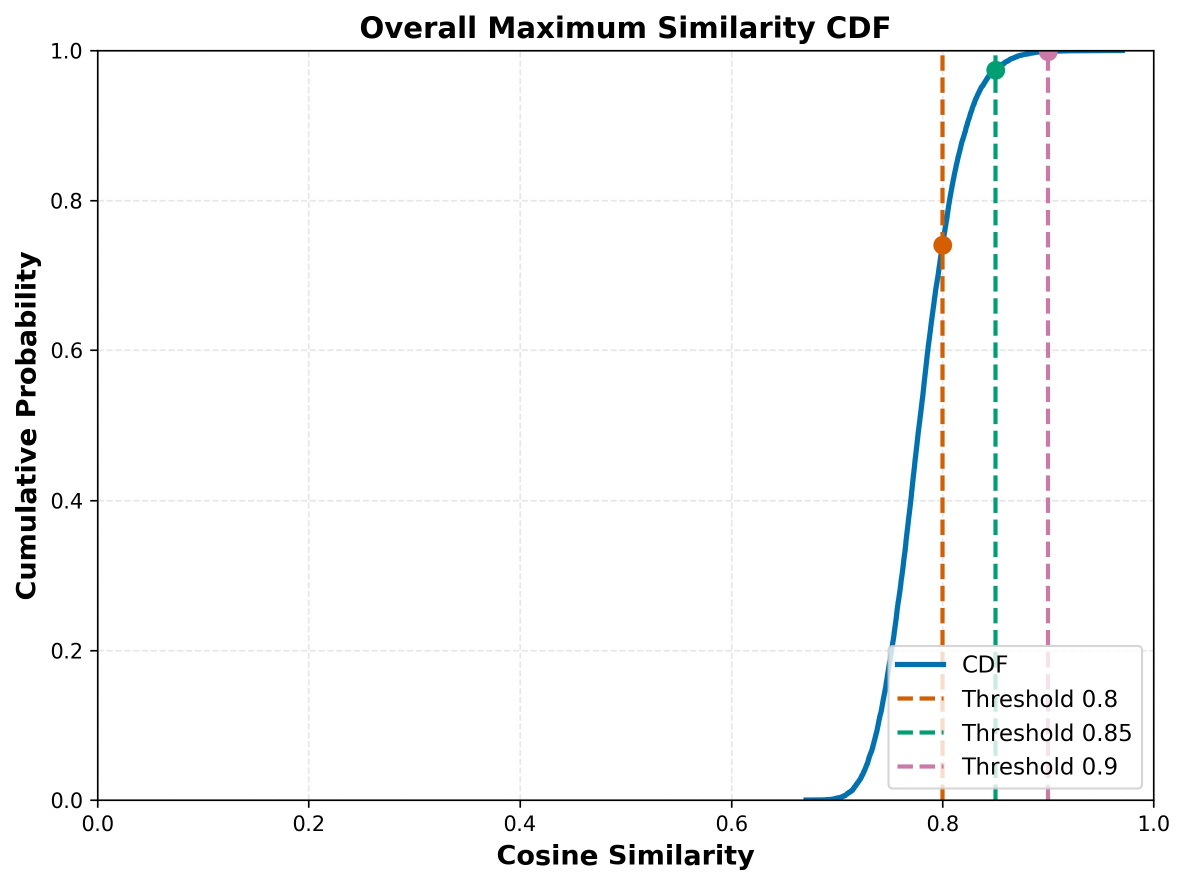}
        \caption{CDF of maximum semantic similarity.}
        \label{fig:contamination_cdf}
    \end{subfigure}
    \hfill
    \begin{subfigure}[b]{0.74\linewidth}
        \centering
        \includegraphics[width=\linewidth]{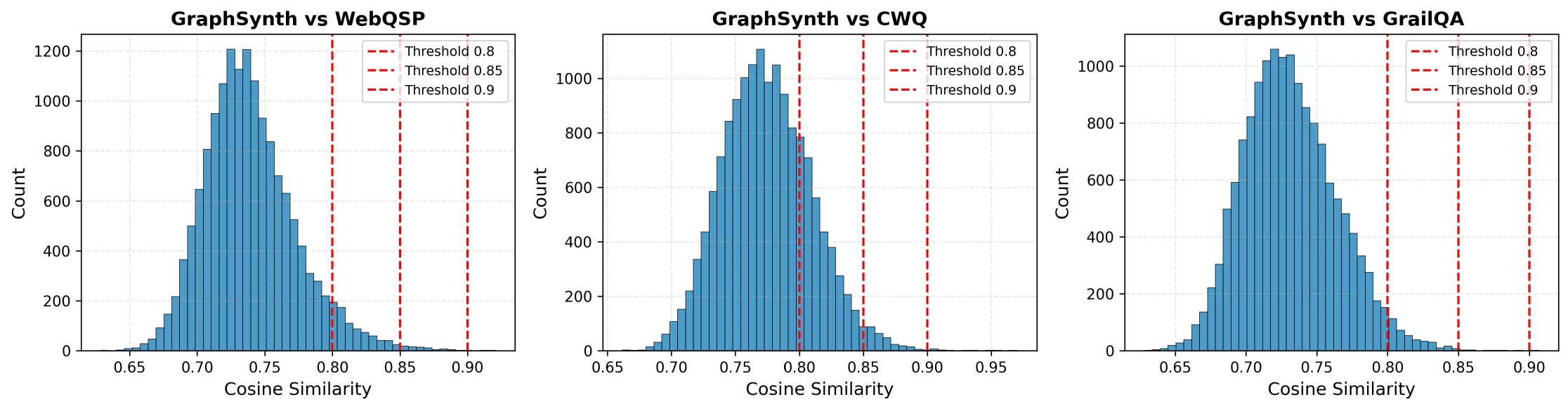}
        \caption{Semantic cosine similarity distribution on each benchmark.}
        \label{fig:contamination_hist}
    \end{subfigure}
    \caption{Data contamination analysis between GraphSynth and downstream benchmarks.}
    \label{fig:contamination}
    \vspace{-14pt}
\end{figure*}

To preclude knowledge leakage in GraphSynth-15k, we compute the maximum cosine similarity between each synthetic question and all test questions across WebQSP, CWQ, and GrailQA using BGE-M3~\citep{chen2024m3}.
At a strict threshold of 0.85, only 2.57\% of GraphSynth questions exhibit high semantic overlap with any test question, and all flagged samples are excluded from the final training corpus.
The elevated overlap rate against CWQ at threshold 0.80 (23.30\%) reflects expected domain proximity rather than leakage, as CWQ encompasses the most diverse relations and broadest domain coverage among the three benchmarks.
Table~\ref{tab:contamination} and Figure~\ref{fig:contamination_cdf} present the full results.

\begin{table}[h]
\centering
\small
\begin{tabular}{lcccc}
\toprule
\textbf{Threshold} & \textbf{Overall} & \textbf{CWQ} & \textbf{WebQSP} & \textbf{GrailQA} \\
\midrule
0.80 & 25.97\% (3960) & 23.30\% (3553) & 5.27\% (804) & 2.67\% (407) \\
0.85 &  2.57\% (392)  &  2.13\% (325)  & 0.56\% (85)  & 0.09\% (13)  \\
0.90 &  0.15\% (23)   &  0.12\% (18)   & 0.03\% (4)   & 0.01\% (1)   \\
\bottomrule
\end{tabular}
\caption{Contamination rates at three similarity thresholds (BGE-M3).}
\label{tab:contamination}
\end{table}

\subsection{Statistics of GraphSynth}
\label{app:stat_graphsynth}

\begin{table}[htbp]
  \centering
  \small
  \setlength{\tabcolsep}{6pt}
  \begin{tabular}{l cccc c c}
    \toprule
    \multirow{2}{*}{\textbf{Structure Type}} & \multicolumn{4}{c}{\textbf{Composition}} & \textbf{Conjunction} & \multirow{2}{*}{\textbf{Total}} \\
    \cmidrule(lr){2-5} \cmidrule(lr){6-6}
    & \textbf{2-hop} & \textbf{3-hop} & \textbf{4-hop} & \textbf{5-hop} & \textbf{2I} & \\
    \midrule
    \textbf{Count} & 1,982 & 4,920 & 1,778 & 576 & 5,599 & 14,855 \\
    \textbf{Ratio (\%)} & 13.34 & 33.12 & 11.97 & 3.88 & 37.69 & 100.00 \\
    \bottomrule
  \end{tabular}
  \caption{Structure distribution of the GraphSynth-15k dataset.}
  \label{tab:graphtrails-e-type}
\end{table}

\begin{wrapfigure}{r}{0.40\textwidth}
    \vspace{-12pt}
    \centering
    \includegraphics[width=\linewidth]{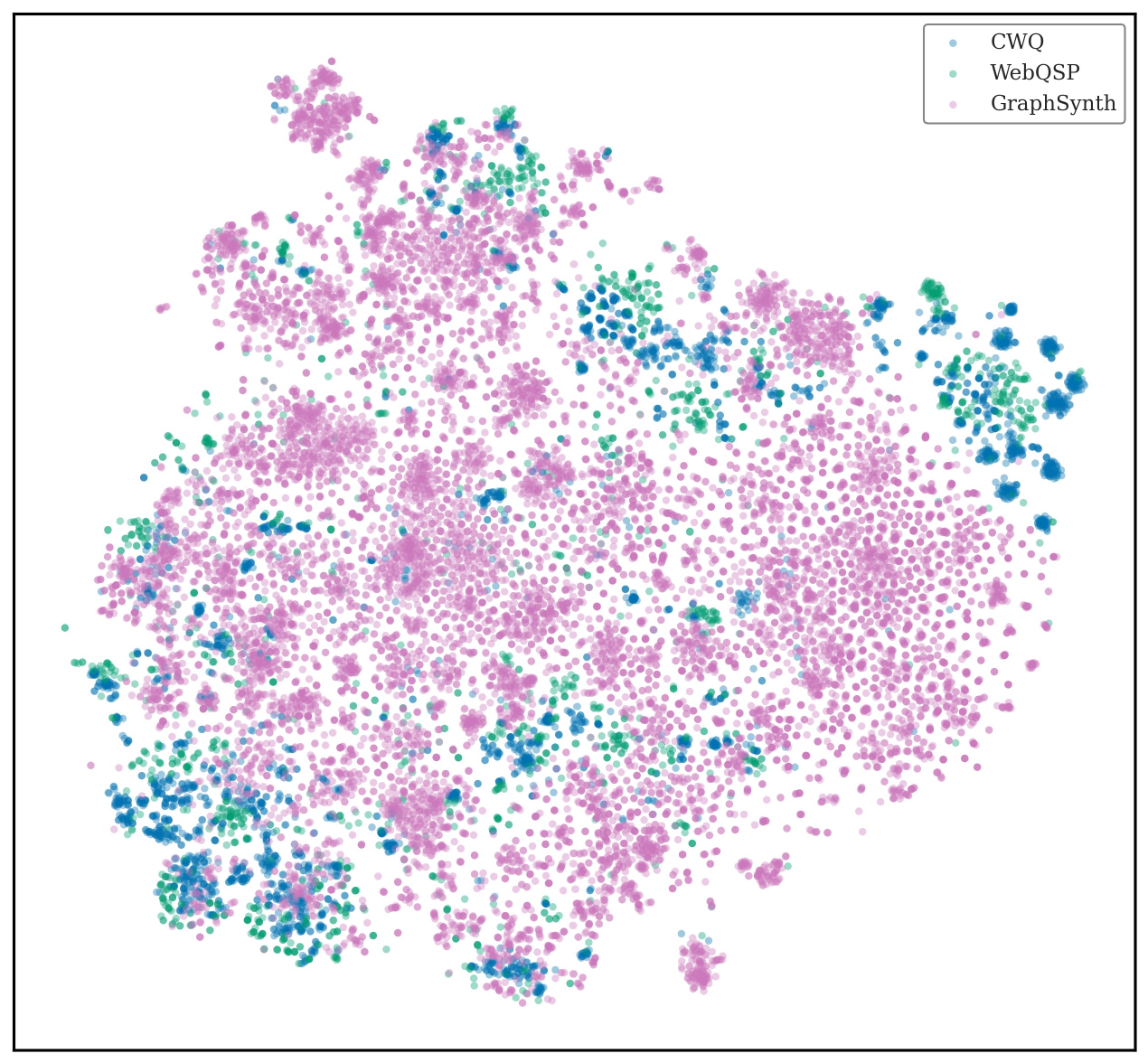}
    \caption{t-SNE visualization of GraphSynth, CWQ, and WebQSP question embeddings, illustrating the broad semantic coverage of the synthesized corpus.}
    \label{fig:tsne}
    \vspace{-10pt}
\end{wrapfigure}

In this section, we characterize GraphSynth-15k along two dimensions to validate its suitability as a training corpus.
Structurally, as summarized in Table~\ref{tab:graphtrails-e-type}, GraphSynth spans five reasoning structure types, covering multi-hop compositions (2- to 5-hop) and conjunctions (2I), with 3-hop and 2I patterns accounting for over 70\% of samples while longer chains ensure exposure to complex reasoning patterns.
Semantically, the synthesized questions are grounded in Freebase relations drawn from diverse topical domains, achieving broad coverage of real-world entity types and relation schemas.
As visualized in Figure~\ref{fig:tsne}, the question embeddings of GraphSynth spread across a wide region of the semantic space, overlapping substantially with both CWQ and WebQSP without concentrating around any single benchmark cluster, confirming that the corpus is neither narrowly specialized nor artificially constrained to a particular domain.
Together, these properties ensure that GraphSynth provides structurally and semantically diverse supervision for Stage 1 training.

\subsection{Structural Contamination Analysis}
\label{app:structural_contamination}

To complement question-level semantic filtering, we compare GraphSynth relation paths with the ground-truth reasoning paths of all 3,531 CWQ test examples. A test example is considered covered if any of its ground-truth relation paths exactly matches a GraphSynth path. This occurs for only 85 examples (2.41\%), indicating low exact structural overlap.

\subsection{Data Construction Cost}
\label{app:construction_cost}

Table~\ref{tab:construction_cost} reports the average per-sample API usage and estimated cost for constructing GraphSynth and GraphRoll using \texttt{GPT-4o-mini}. All values are averaged across samples in each dataset.

\begin{table}[h]
\centering
\small
\setlength{\tabcolsep}{6pt}
\renewcommand{\arraystretch}{1.15}
\begin{tabular}{lrrrr}
\toprule
\textbf{Dataset} & \textbf{API Calls} & \textbf{Input Tokens} & \textbf{Output Tokens} & \textbf{Cost} \\
\midrule
GraphSynth & 6.673 & 11,533.56 & 796.53 & \textbf{\$0.0110} \\
GraphRoll  & 5.680 &  9,296.93 & 553.87 & \textbf{\$0.0086} \\
\bottomrule
\end{tabular}
\caption{Average per-sample API usage and cost for GraphSynth and GraphRoll data construction.}
\label{tab:construction_cost}
\end{table}

\section{Datasets}
\label{app:datasets}
In this paper, we conduct experiments on three widely used 
KGQA datasets, and we also construct our own GraphWalkerBench to further study the agent's capability to generalize to unseen structures: 

\textbf{WebQSP.} WebQSP is a foundational KGQA benchmark designed to evaluate a model’s ability to answer simple, fact-based questions that typically require retrieving a single fact from the knowledge graph. The dataset provides full SPARQL queries for its questions, which are executed against Freebase.

\textbf{CWQ.} CWQ extends the complexity beyond WebQSP by introducing questions that require compositional reasoning. These questions often involve multiple constraints, conjunctions, and superlatives, necessitating multi-hop or multi-relation inference paths on the Freebase knowledge graph.
The dataset is annotated with complex SPARQL queries that reflect these reasoning structures.

\textbf{GrailQA.} GrailQA is a large-scale dataset built on Freebase that is designed to evaluate the generalization capabilities of KGQA models across three distinct levels: i.i.d., compositional, and zero-shot. This structure allows for a fine-grained analysis of a model’s ability to handle previously seen patterns (i.i.d.), new combinations of seen patterns (compositional), and entirely new domains and relations (zero-shot).
Significantly, the zero-shot split in GrailQA evaluates the model’s ability to generalize to unseen query compositions.

\textbf{GraphWalkerBench.}
GraphWalkerBench consists of 1766 queries covering a diverse set of reasoning 
structures, including 736 multi-hop compositions and 1030 conjunction patterns. 
As detailed in Table~\ref{tab:graphwalkerbench}, it specifically includes 
single-path composition queries (3-hop, 4-hop, and 5-hop) and multi-path 
conjunction queries (2I, IP, and PI), providing a balanced testbed for 
evaluating both depth and compositional complexity.

\begin{table}[htbp]
  \centering
  \small
  \setlength{\tabcolsep}{8pt}
  \begin{tabular}{l ccc ccc c}
    \toprule
    \multirow{2}{*}{\textbf{Structure Type}} & \multicolumn{3}{c}{\textbf{Composition}} & \multicolumn{3}{c}{\textbf{Conjunction}} & \multirow{2}{*}{\textbf{Total}} \\
    \cmidrule(lr){2-4} \cmidrule(lr){5-7}
    & \textbf{3-hop} & \textbf{4-hop} & \textbf{5-hop} & \textbf{2I} & \textbf{IP} & \textbf{PI} & \\
    \midrule
    \textbf{Count} & 347 & 209 & 180 & 344 & 337 & 349 & 1766 \\
    \bottomrule
  \end{tabular}
  \caption{Structure distribution of GraphWalkerBench.}
  \label{tab:graphwalkerbench}
\end{table}

Beyond structural depth and composition type, GraphWalkerBench covers 909 unique KG relations across diverse domains, including film, location, people, books, fictional universes, sports, television, and organizations. We conduct a stratified manual audit across reasoning types to verify that each question is answerable from its annotated KG path or paths, that the annotated answer is consistent with those paths, and that the question does not directly leak the answer. Table~\ref{tab:graphwalkerbench_audit} presents representative audited examples.

\begin{table*}[htbp]
\centering
\scriptsize
\setlength{\tabcolsep}{4pt}
\renewcommand{\arraystretch}{1.15}
\begin{tabularx}{\textwidth}{l>{\raggedright\arraybackslash}X>{\raggedright\arraybackslash}X}
\colortoprule
\rowcolor[HTML]{D9DAE3}
\textbf{Type} & \textbf{Annotated path(s)} & \textbf{Question} \\
\colormidrule
2I & (1) Korean War $\rightarrow$ films $\rightarrow$ MASH; 

(2) Ring Lardner, Jr. $\rightarrow$ writer.film $\rightarrow$ MASH & What film about the Korean War was written by Ring Lardner, Jr.? \\
\midrule
PI & (1) Czech Republic $\rightarrow$ Central Europe $\rightarrow$ Europe; 

(2) Continental Celtic languages $\rightarrow$ geographic distribution $\rightarrow$ Europe & What continent that includes the Czech Republic is also the geographic home of the Continental Celtic languages? \\
\midrule
4-hop & Classic rock $\rightarrow$ Rock music $\rightarrow$ Rock and Roll Hall of Fame $\rightarrow$ I. M. Pei $\rightarrow$ Pei Cobb Freed \& Partners & What architectural firm was a partner of the architect who designed a hall of fame for a parent genre of classic rock? \\
\midrule
5-hop & Rotokas Language $\rightarrow$ Papuan languages $\rightarrow$ New Guinea $\rightarrow$ Malay Archipelago $\rightarrow$ Lost Continent $\rightarrow$ Angelo Francesco Lavagnino & Who composed the music for a film about the island group home to the Rotokas language family? \\
\bottomrule
\end{tabularx}
\caption{Representative manually audited examples from GraphWalkerBench.}
\label{tab:graphwalkerbench_audit}
\end{table*}

\begin{figure}[htbp]
    \centering
    \begin{minipage}[t]{0.48\linewidth}
        \centering
        \includegraphics[width=\linewidth]{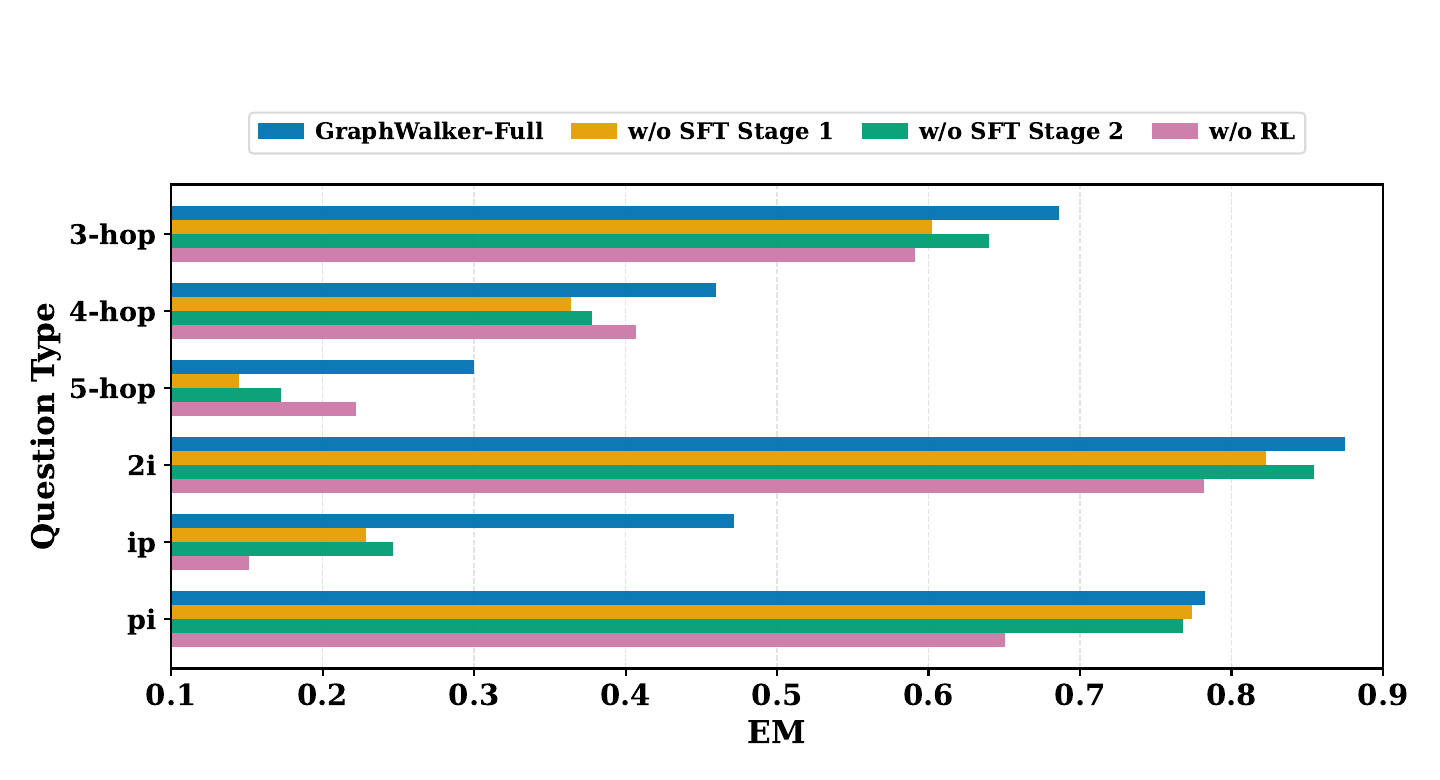}
    \end{minipage}
    \hfill
    \begin{minipage}[t]{0.48\linewidth}
        \centering
        \includegraphics[width=\linewidth]{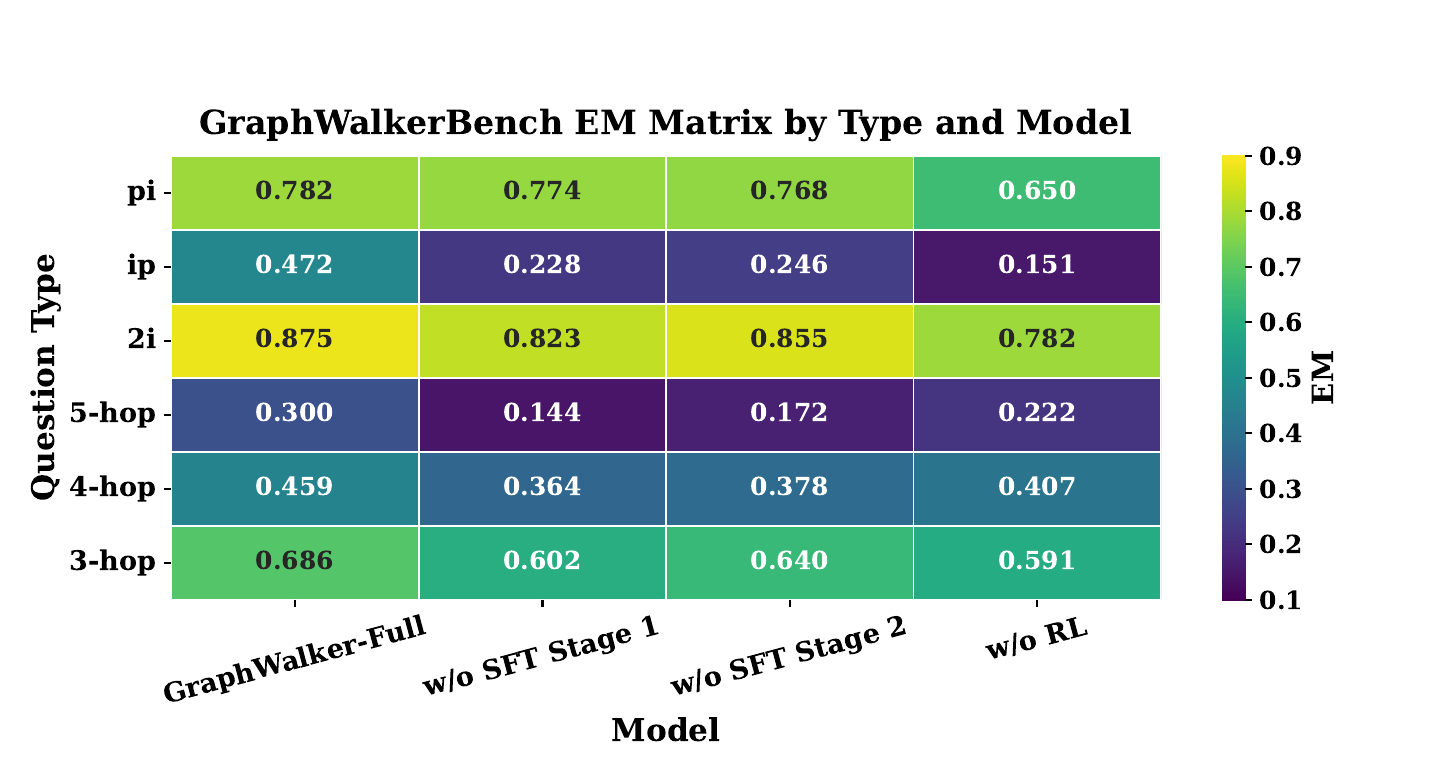}
    \end{minipage}
    \caption{Fine-grained EM performance on GraphWalkerBench across question 
    types and ablation variants. \textbf{Left}: grouped bar chart. 
    \textbf{Right}: heatmap. IP and PI are unseen conjunction structures 
    not present in GraphSynth training data.}
    \label{fig:gwbench_analysis}
\end{figure}

In Figure~\ref{fig:gwbench_analysis}, we present a fine-grained breakdown of EM performance across question types on GraphWalkerBench. 
\textbf{GraphWalker-Full} achieves the highest EM across nearly all types, with particularly strong generalization to unseen conjunction structures (IP and PI), confirming that GraphSynth is the primary driver of structural 
generalization. Removing either SFT stage or RL consistently degrades performance, with severe drops observed on IP queries and deep multi-hop compositions, corroborating the findings in Table~\ref{tab:zeroshot}.

\section{Baselines}
\label{app:baselines}

We compare GraphWalker with a comprehensive set of baselines.
Following previous work, the Freebase dump is acquired from \url{https://developers.google.com/freebase?hl=en}. 
We deploy Freebase with Virtuoso. 
All the baselines are evaluated on Freebase database.

\textbf{IO-Prompt.}
IO-Prompt evaluates the intrinsic parametric knowledge of LLMs by directly feeding the natural language question and requesting a final answer. No external knowledge graph or intermediate reasoning steps are involved. This setting serves as a baseline for the model's zero-shot performance without grounding.

\textbf{Reasoning on Graphs (RoG).}
RoG \citep{luo2023reasoning} decouples KGQA into planning and retrieval-reasoning. It first fine-tunes an LLM to generate relation paths as reasoning plans, which are then used to retrieve relevant subgraphs from the KG. Finally, a reader module reasons over the retrieved subgraphs to answer the question. Unlike GraphWalker, RoG's planning is static and does not dynamically interact with the KG environment during the planning phase.

\textbf{Think-on-Graph (ToG).} 
ToG \citep{sun2023think} integrates structured knowledge from a knowledge graph into the LLM’s reasoning process, representing the methods using LLM generation to perform knowledge retrieval. It utilizes KG-based retrieval to enrich the model’s input, allowing it to leverage explicit entity-relation structures for improved factual accuracy. At each step, it uses the LLM to decide whether to continue retrieving or terminate the process.

\textbf{Think-on-Graph 2.0 (ToG-2.0).}
ToG-2 \citep{ma2025thinkongraph20deepfaithful} introduces a tightly-coupled hybrid RAG framework that iteratively integrates structured KGs and unstructured document contexts. It alternates between knowledge-guided text retrieval and context-enhanced graph search to gather in-depth heterogeneous clues. At each step, the LLM evaluates the retrieved knowledge to decide whether to continue exploring or generate the final answer.

\textbf{Generate-on-Graph (GoG).}
GoG \citep{xu2024generate} addresses KGQA under incomplete knowledge graphs via a training-free Thinking-Searching-Generating framework, where the LLM acts simultaneously as an agent to explore the KG and as a knowledge source to generate missing factual triples on demand. When retrieved subgraphs lack sufficient information, GoG invokes a Generate Action that synthesizes new triples grounded in the explored context and verifies them before reasoning, enabling the model to bridge structural gaps in the KG using its parametric knowledge.

\textbf{KG-Agent.}
KG-Agent \citep{jiang2025kg} adopts a tool-learning paradigm. It fine-tunes a smaller LLM (e.g., Llama-2-7B) to act as an autonomous agent that synthesizes executable tool calls (e.g., logical operations, KG queries). The model is trained on a synthesized dataset derived from functional programs, enabling it to decompose complex questions into sequences of tool executions.

\textbf{KBQA-o1.}
KBQA-o1 \citep{luo2025kbqa} proposes an agentic KBQA framework that combines a ReAct-based agent process with Monte Carlo Tree Search (MCTS) for heuristic KB environment exploration, where a policy model guides stepwise logical form generation and a reward model scores complete reasoning trajectories to balance search effectiveness and efficiency. To reduce reliance on human annotation, it further employs MCTS-driven exploration over unlabeled questions to auto-generate high-quality training data for incremental fine-tuning of both models.

\textbf{KG-R1.}
KG-R1~\citep{song2025efficient} is a state-of-the-art RL framework for KGQA that trains a lightweight LLM as a reasoning agent, optimizing its retrieval and reasoning policy end-to-end via turn-level rewards to navigate the KG efficiently. Unlike GraphWalker, KG-R1 does not employ a decoupled exploration pre-training stage. For a fair comparison, we equip KG-R1 with \textit{the same} four tools as specified in their original paper.

\section{Further Analysis}
\subsection{Fine-grained RL Effect by Reasoning Type}
\label{app:rl_radar_bytype}

\begin{figure}[h]
  \centering
  \includegraphics[width=0.62\linewidth]{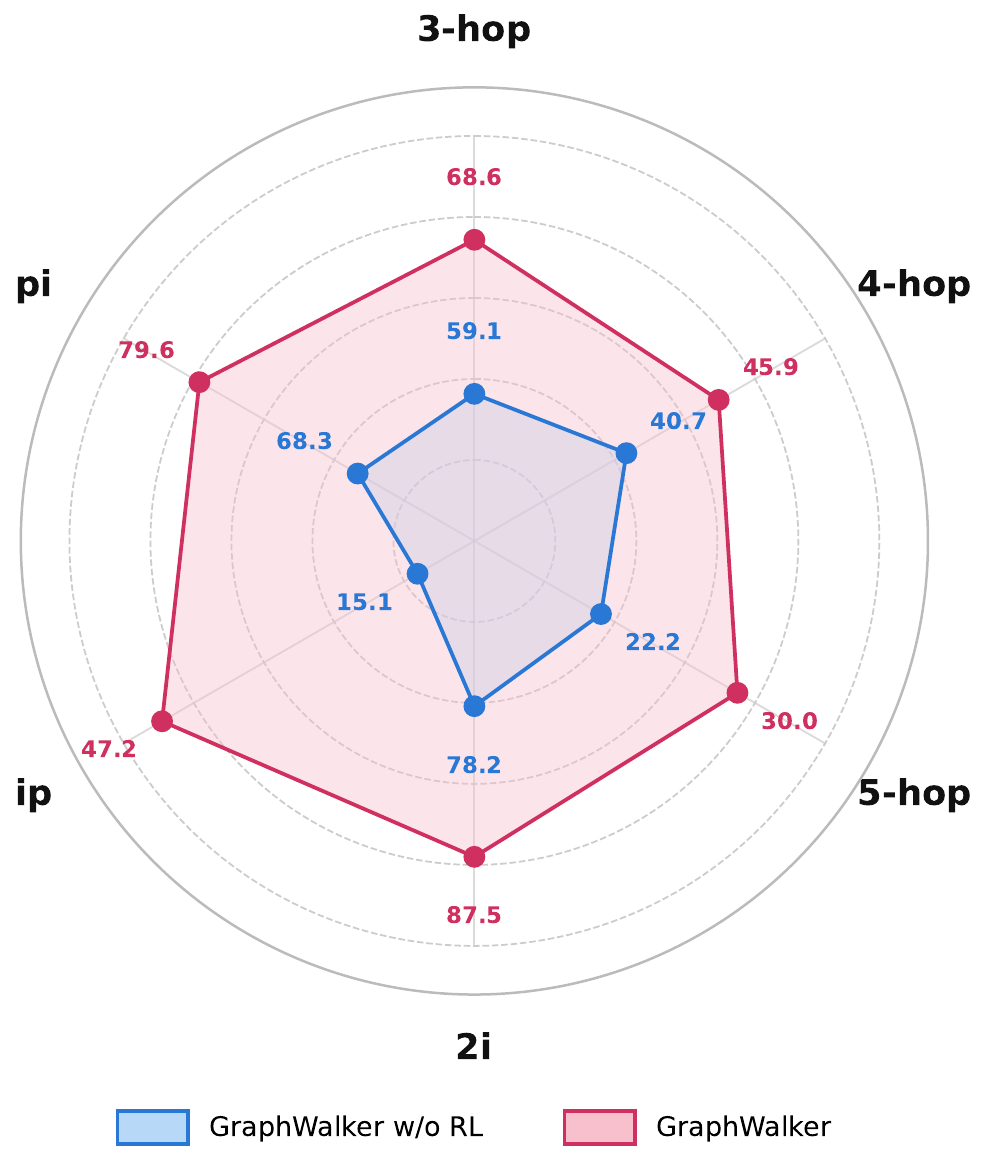}
  \caption{Fine-grained effect of RL across GraphWalkerBench reasoning structures. Each axis is independently scaled.}
  \label{fig:rl_radar_bytype}
\end{figure}

Figure~\ref{fig:rl_radar_bytype} shows the per-reasoning-type EM comparison between \textit{GraphWalker} and \textit{GraphWalker w/o RL} on GraphWalkerBench. RL improves performance across all six reasoning structures, with the largest absolute gain on IP (+32.05 EM), where the model must combine path exploration with intersection-style constraints.

\subsection{Ablation of LLM-based Relation Reranking}
The relation reranking module serves a critical role in our framework: 
since BM25 retrieval can return a large number of candidate relations, 
feeding them directly into the LLM would result in excessively long contexts 
and distract the agent with irrelevant information. 
To investigate its effectiveness and determine the optimal candidate threshold $k$, we conduct an ablation study on CWQ and WebQSP using \texttt{Qwen2.5-7B-Instruct} as 
the filter model, with results summarized in Table~\ref{tab:reranking_ablation}.

\begin{table}[htbp]
\centering
\small
\setlength{\tabcolsep}{16pt}
\renewcommand{\arraystretch}{1.1}
\begin{tabular}{l|cc|cc}
\toprule
\multicolumn{1}{c|}{\textbf{Method}}
& \multicolumn{2}{c|}{\textbf{CWQ}}  
& \multicolumn{2}{c}{\textbf{WebQSP}} \\
& \textbf{EM} & \textbf{F1}  
& \textbf{EM} & \textbf{F1} \\
\midrule
\colormidrule
\rowcolor{blue!8}
\textbf{w/ Reranking Top-20}      
& \textbf{79.6} & \underline{74.2} 
& \textbf{91.5} & \textbf{88.6} \\
\colormidrule
\rowcolor{gray!15}
\multicolumn{5}{c}{\textit{Ablations}} \\
\colormidrule
w/ Reranking Top-10    & 78.4 & 73.5 & 89.4 & 86.1 \\
w/ Reranking Top-15    & \underline{78.7} & \textbf{74.6} & \underline{89.8} & \underline{86.5} \\
w/ Reranking Top-30    & 76.9 & 72.0 & 87.0 & 84.4 \\
w/o Reranking          & 74.8 & 70.1 & 87.0 & 82.2 \\
\bottomrule
\end{tabular}
\caption{LLM-based Relation Reranking Ablation Study. The best and second-best results are shown in \textbf{bold} and \underline{underlined}, respectively.}
\label{tab:reranking_ablation}
\end{table}

Notably, even without reranking (\textbf{w/o Reranking}), GraphWalker remains 
competitive, achieving 74.8\% EM on CWQ and 87.0\% EM on WebQSP. This 
demonstrates the robustness of our interaction framework even under degraded 
retrieval conditions, where the agent relies solely on BM25-retrieved relations 
without any LLM-based filtering.

Nevertheless, incorporating reranking consistently improves performance, and 
the choice of $k$ reveals a clear trend. Increasing $k$ from 10 to 20 yields 
steady gains, as a larger candidate pool improves the recall of crucial 
relations that may be missed by lower-$k$ settings. The \textbf{Top-20} setting 
achieves the best overall balance, attaining the highest EM on both CWQ 
(79.6\%) and WebQSP (91.5\%). However, further expanding to \textbf{Top-30} 
reverses this trend: although retaining more candidates reduces false negatives, 
it simultaneously introduces excessive noise into the context, distracting the 
agent during trajectory exploration and ultimately degrading performance. 
Consequently, we adopt Top-20 as the default threshold for our framework.

\subsection{Case Study}

\label{app:case_study}
In this section, we present a representative case comparing three model variants: \textbf{GraphWalker-SFT-GraphSynth}, \textbf{GraphWalker-SFT-GraphRoll}, and \textbf{GraphWalker-SFT-RL}. We analyze their reasoning trajectories on one question from CWQ: \textit{``Where is the home stadium of the team who won the 1946 World Series championship?''}
As shown in Appendix~\ref{app:four_cases}, \textbf{GraphWalker-SFT-GraphSynth} successfully retrieves the correct triple \texttt{[St. Louis Cardinals, arena\_stadium, Busch Stadium]} at Turn 3, and even further verifies it via an additional query at Turn 4, explicitly confirming \textit{``Busch Stadium is the home venue for the St. Louis Cardinals''} in its reasoning, yet it outputs \texttt{[``St. Louis Cardinals'']} instead of extracting the stadium name---this reveals that Stage 1 SFT (on GraphSynth) teaches KG navigation but not answer extraction.
\textbf{GraphWalker-SFT-GraphRoll} identifies the champion St. Louis Cardinals at Turn 2 and observes \texttt{Sportsman's Park} in the event locations, but directly outputs this historical venue \texttt{[``Sportsman's Park'']} without further exploring the team's current stadium relation---this shows that Stage 2 SFT (on GraphRoll) alone, trained on short expert trajectories, lacks the multi-hop exploration capacity needed for deeper reasoning paths.
\textbf{GraphWalker-SFT-RL} correctly identifies the champion at Turn 2, then directly queries the St. Louis Cardinals' stadium relation and outputs \texttt{[``Busch Stadium'']} as the final answer (EM=1.0), demonstrating that RL consolidates the exploration capacity of Stage 1 and the reflection capability of Stage 2 into a more efficient and accurate reasoning strategy.

\subsection{Complete Reasoning Trajectories}
\label{app:four_cases}

We present the complete reasoning trajectories of all four models. Each trajectory is displayed in a table where different message types are color-coded: \colorbox{blue!10}{thinking}, \colorbox{green!10}{KG queries}, \colorbox{yellow!10}{information feedback}, \colorbox{red!10}{errors}, and final answers.

\subsubsection{GraphWalker-SFT-GraphSynth}

{\scriptsize
\setlength{\arrayrulewidth}{0.3pt}
\setlength{\tabcolsep}{8pt}
\renewcommand{\arraystretch}{1.3}
\begin{longtable}{>{\raggedright\arraybackslash}p{0.97\textwidth}}
\colortoprule
\rowcolor{gray!20}
\textbf{Question:} Where is the home stadium of the team who won the 1946 World Series championship? \\
\rowcolor{gray!20}
\textbf{Initial Entities:} 1946 World Series \\
\colormidrule
\rowcolor{blue!10}
\texttt{<think>} Since I need to determine the home stadium of the team that won the 1946 World Series, I will start by querying the relations for the entity "1946 World Series". This will help me identify relevant connections to the winning team and its stadium. Gathering these relations is essential for finding the necessary information to answer the question. \texttt{</think>} \\
\rowcolor{green!10}
\texttt{<kg-query>} get\_relations("1946 World Series") \texttt{</kg-query>} \\
\rowcolor{yellow!10}
\texttt{<information>} sports.sports\_championship\_event.champion, sports.sports\_championship\_event.result, sports.sports\_team.championships, time.event.locations, ... \texttt{</information>} \\
\rowcolor{blue!10}
\texttt{<think>} The relations obtained for "1946 World Series" include "sports.sports\_championship\_event.champion," which is crucial for identifying the winning team, and "time.event.locations," which may provide information about the home stadium. By querying the triples with these relations, I can gather detailed information about the champion team and its associated stadium, which is necessary to answer the question accurately. \texttt{</think>} \\
\rowcolor{green!10}
\texttt{<kg-query>} get\_triples("1946 World Series", ["sports.sports\_championship\_event.champion", "sports.sports\_championship\_event.result", "sports.sports\_team.championships", "time.event.locations", ...]) \texttt{</kg-query>} \\
\rowcolor{yellow!10}
\texttt{<information>} [1946 World Series, sports.sports\_championship\_event.champion, St. Louis Cardinals], [1946 World Series, sports.sports\_championship\_event.result, 4-3], [1946 World Series, sports.sports\_championship\_event.championship, World Series], [St. Louis Cardinals, sports.sports\_team.championships, 1946 World Series] \texttt{</information>} \\
\rowcolor{blue!10}
\texttt{<think>} I learned that the champion of the 1946 World Series is the St. Louis Cardinals. To find the home stadium of this team, I need to query the relations for "St. Louis Cardinals," as this will provide insights into their stadium and other relevant details. This step is essential to connect the team with its home venue, which is necessary to answer the question accurately. \texttt{</think>} \\
\rowcolor{green!10}
\texttt{<kg-query>} get\_relations("St. Louis Cardinals") \texttt{</kg-query>} \\
\rowcolor{yellow!10}
\texttt{<information>} sports.sports\_team.venue, sports.sports\_team.arena\_stadium, sports.sports\_team.location, ... \texttt{</information>} \\
\rowcolor{blue!10}
\texttt{<think>} The relations obtained for "St. Louis Cardinals" include "sports.sports\_team.venue" and "sports.sports\_team.arena\_stadium," which are directly relevant for identifying the team's home stadium. By querying the triples using these relations, I can gather the specific information needed to answer the question about where the St. Louis Cardinals play their home games. \texttt{</think>} \\
\rowcolor{green!10}
\texttt{<kg-query>} get\_triples("St. Louis Cardinals", ["sports.sports\_team.venue", "sports.sports\_team.arena\_stadium", "sports.sports\_team.location", ...]) \texttt{</kg-query>} \\
\rowcolor{yellow!10}
\texttt{<information>} [St. Louis Cardinals, sports.sports\_team.arena\_stadium, Busch Stadium], [St. Louis Cardinals, sports.sports\_team.arena\_stadium, Roger Dean Stadium], [St. Louis Cardinals, sports.sports\_team.location, St. Louis], ... \texttt{</information>} \\
\rowcolor{blue!10}
\texttt{<think>} The last observation revealed that the St. Louis Cardinals have multiple venues listed, including "Busch Stadium." However, to confirm the current home stadium, I need to further investigate the relations associated with "Busch Stadium." This will help clarify if it is indeed the primary venue used by the team for their home games, which is crucial for answering the question accurately. \texttt{</think>} \\
\rowcolor{green!10}
\texttt{<kg-query>} get\_relations("Busch Stadium") \texttt{</kg-query>} \\
\rowcolor{yellow!10}
\texttt{<information>} sports.sports\_facility.home\_venue\_for, sports.sports\_facility.teams, sports.sports\_team.arena\_stadium, ... \texttt{</information>} \\
\rowcolor{blue!10}
\texttt{<think>} The relations obtained for "Busch Stadium" include "sports.sports\_facility.home\_venue\_for" and "sports.sports\_facility.teams," which are critical for confirming whether it serves as the home stadium for the St. Louis Cardinals. By querying the triples with these relations, I can verify the connection between the stadium and the team, ensuring that I accurately identify the home stadium for the team that won the 1946 World Series. \texttt{</think>} \\
\rowcolor{green!10}
\texttt{<kg-query>} get\_triples("Busch Stadium", ["sports.sports\_facility.home\_venue\_for", "sports.sports\_facility.teams", "sports.sports\_team.arena\_stadium", ...]) \texttt{</kg-query>} \\
\rowcolor{yellow!10}
\texttt{<information>} [Busch Stadium, sports.sports\_facility.teams, St. Louis Cardinals], [St. Louis Cardinals, sports.sports\_team.arena\_stadium, Busch Stadium], ... \texttt{</information>} \\
\rowcolor{blue!10}
\texttt{<think>} The last observation confirmed that Busch Stadium is indeed the home venue for the St. Louis Cardinals, as indicated by the triple showing "[Busch Stadium, sports.sports\_facility.teams, St. Louis Cardinals]." Since the question asks for the home stadium of the team that won the 1946 World Series, I can confidently conclude that Busch Stadium is the correct answer. Thus, I will provide the answer now. \texttt{</think>} \\
\rowcolor{red!10}
\texttt{<answer>} ["St. Louis Cardinals"] \texttt{</answer>} \textbf{(EM=0.0 - outputs team name instead of stadium)} \\
\colorbottomrule
\end{longtable}
}

\newpage
\subsubsection{GraphWalker-SFT-GraphRoll}

{\scriptsize
\setlength{\arrayrulewidth}{0.3pt}
\setlength{\tabcolsep}{8pt}
\renewcommand{\arraystretch}{1.3}
\begin{longtable}{>{\raggedright\arraybackslash}p{0.97\textwidth}}
\colortoprule
\rowcolor{gray!20}
\textbf{Question:} Where is the home stadium of the team who won the 1946 World Series championship? \\
\rowcolor{gray!20}
\textbf{Initial Entities:} 1946 World Series \\
\colormidrule
\rowcolor{blue!10}
\texttt{<think>} To find the home stadium of the team that won the 1946 World Series championship, I will start by querying the relations of the entity "1946 World Series". This should provide information about the winning team and potentially its home stadium. After retrieving the relations, I will look for relevant details regarding the championship and the winning team. \texttt{</think>} \\
\rowcolor{green!10}
\texttt{<kg-query>} get\_relations("1946 World Series") \texttt{</kg-query>} \\
\rowcolor{yellow!10}
\texttt{<information>} sports.sports\_championship\_event.champion, sports.sports\_championship\_event.result, sports.sports\_team.championships, time.event.locations, ... \texttt{</information>} \\
\rowcolor{blue!10}
\texttt{<think>} The relations retrieved for the "1946 World Series" include information about the champion, result, championships associated with the team, locations, start date, end date, and runner-up. To find the home stadium of the winning team, I will focus on the "champion" relation first, as it should indicate which team won the championship. Once I identify the champion, I can then look for its home stadium. I will proceed to retrieve the triples for the "1946 World Series" using the top relevant relations. \texttt{</think>} \\
\rowcolor{green!10}
\texttt{<kg-query>} get\_triples("1946 World Series", ["sports.sports\_championship\_event.champion", "sports.sports\_championship\_event.result", "sports.sports\_team.championships", "time.event.locations"]) \texttt{</kg-query>} \\
\rowcolor{yellow!10}
\texttt{<information>} [1946 World Series, sports.sports\_championship\_event.champion, St. Louis Cardinals], [1946 World Series, sports.sports\_championship\_event.result, 4-3], [St. Louis Cardinals, sports.sports\_team.championships, 1946 World Series], [1946 World Series, time.event.locations, Fenway Park], [1946 World Series, time.event.locations, Sportsman's Park] \texttt{</information>} \\
\rowcolor{blue!10}
\texttt{<think>} The champion is St. Louis Cardinals. The event locations include Sportsman's Park, which is in St. Louis. Since the question asks for the home stadium of the winning team, and Sportsman's Park appears in the event locations, I will conclude this is the home stadium. \texttt{</think>} \\
\rowcolor{red!10}
\texttt{<answer>} ["Sportsman's Park"] \texttt{</answer>} \textbf{(EM=0.0 - stops after 2 turns without exploring team's stadium relation)} \\
\colorbottomrule
\end{longtable}
}

\subsubsection{GraphWalker-SFT-RL}

{\scriptsize
\setlength{\arrayrulewidth}{0.3pt}
\setlength{\tabcolsep}{8pt}
\renewcommand{\arraystretch}{1.3}
\begin{longtable}{>{\raggedright\arraybackslash}p{0.97\textwidth}}
\colortoprule
\rowcolor{gray!20}
\textbf{Question:} Where is the home stadium of the team who won the 1946 World Series championship? \\
\rowcolor{gray!20}
\textbf{Initial Entities:} 1946 World Series \\
\colormidrule
\rowcolor{blue!10}
\texttt{<think>} To find the home stadium of the team that won the 1946 World Series championship, I will start by querying the relations of the "1946 World Series" entity. This should provide information about the winning team and potentially lead to details about their home stadium. \texttt{</think>} \\
\rowcolor{green!10}
\texttt{<kg-query>} get\_relations("1946 World Series") \texttt{</kg-query>} \\
\rowcolor{yellow!10}
\texttt{<information>} sports.sports\_championship\_event.champion, sports.sports\_championship\_event.result, sports.sports\_championship\_event.runner\_up, ... \texttt{</information>} \\
\rowcolor{blue!10}
\texttt{<think>} The relations returned for the "1946 World Series" include several relevant ones, particularly "champion," which likely indicates the winning team. I will also consider "runner\_up" as it may provide context about the teams involved. To gather more detailed information, I will query the triples for the "1946 World Series" using these relations. \texttt{</think>} \\
\rowcolor{green!10}
\texttt{<kg-query>} get\_triples("1946 World Series", ["sports.sports\_championship\_event.champion", "sports.sports\_championship\_event.runner\_up", "sports.sports\_championship\_event.result"]) \texttt{</kg-query>} \\
\rowcolor{yellow!10}
\texttt{<information>} [1946 World Series, sports.sports\_championship\_event.champion, St. Louis Cardinals], [1946 World Series, sports.sports\_championship\_event.runner\_up, Boston Red Sox], [1946 World Series, sports.sports\_championship\_event.result, 4-3], [1946 World Series, sports.sports\_championship\_event.championship, World Series] \texttt{</information>} \\
\rowcolor{blue!10}
\texttt{<think>} The champion of the 1946 World Series is the St. Louis Cardinals, and the runner-up is the Boston Red Sox. To find the home stadium of the St. Louis Cardinals, I will now query the relations for the "St. Louis Cardinals" entity. This should help me identify the stadium associated with this team. \texttt{</think>} \\
\rowcolor{green!10}
\texttt{<kg-query>} get\_relations("St. Louis Cardinals") \texttt{</kg-query>} \\
\rowcolor{yellow!10}
\texttt{<information>} sports.sports\_team.venue, sports.sports\_team.arena\_stadium, sports.sports\_team.location, ... \texttt{</information>} \\
\rowcolor{blue!10}
\texttt{<think>} The relations for the "St. Louis Cardinals" include "venue" and "arena\_stadium," both of which are likely to provide information about the home stadium of the team. I will query the triples for the "St. Louis Cardinals" using these relations to find the specific stadium. \texttt{</think>} \\
\rowcolor{green!10}
\texttt{<kg-query>} get\_triples("St. Louis Cardinals", ["sports.sports\_team.venue", "sports.sports\_team.arena\_stadium", "sports.sports\_team.location"]) \texttt{</kg-query>} \\
\rowcolor{yellow!10}
\texttt{<information>} [St. Louis Cardinals, sports.sports\_team.arena\_stadium, Busch Stadium], [St. Louis Cardinals, sports.sports\_team.arena\_stadium, Roger Dean Stadium], [St. Louis Cardinals, sports.sports\_team.location, St. Louis], ... \texttt{</information>} \\
\rowcolor{blue!10}
\texttt{<think>} The query results indicate that the St. Louis Cardinals have multiple stadiums associated with them, including Busch Stadium and Sportsman's Park. Since Busch Stadium is mentioned multiple times and is the most recent name used, it is likely the primary home stadium of the team. Therefore, I will provide Busch Stadium as the answer. \texttt{</think>} \\
\hline
\rowcolor{green!20}
\texttt{<answer>} ["Busch Stadium"] \texttt{</answer>} \textbf{(EM=1.0 - correct answer with shortest path)} \\
\colorbottomrule
\end{longtable}
}

\section{KG deployment}
In this section, we describe the KG deployment infrastructure, including SPARQL query tools, constrained random walk, and optimizations.
We deploy Virtuoso 7.2.5 with Freebase. 
Entity types and names are specified via \texttt{type.object.type} and \texttt{type.object.name} with language tags. CVT nodes without readable names are flattened into two-hop relations (details are shown in section~\ref{sec:get_triples}).

\subsection{Entity Resolution}
\label{sec:entity_resolution}

Entity names are resolved to MIDs via exact match on \texttt{type.object.name@en}. Multiple candidates are disambiguated by selecting the entity with most types. A case-insensitive fallback is applied if needed. Results are cached. Listing~\ref{lst:entity_query} details this process.

\begin{lstlisting}[caption={Entity resolution query.},label=lst:entity_query,float=ht]
PREFIX ns: <http://rdf.freebase.com/ns/>
SELECT DISTINCT ?entity ?type WHERE {
  ?entity ns:type.object.name "Barack Obama"@en .
  ?entity ns:type.object.type ?type .
} LIMIT 100
\end{lstlisting}

\subsection{\texttt{get\_relations}}
\label{sec:get_relations}

\texttt{get\_relations(entity)} returns relation IDs. We split outgoing/incoming queries to avoid timeout, merge results, apply whitelist filtering, and optionally rank by BM25 similarity to return top-\textbf{k}. CVT flatten relations are appended. See Listing~\ref{lst:relations_query}.

\begin{lstlisting}[caption={Outgoing/incoming relation queries.},label=lst:relations_query,float=ht]
-- Outgoing
PREFIX ns: <http://rdf.freebase.com/ns/>
SELECT DISTINCT ?relation WHERE {
  ns:m.02mjmr ?relation ?tail .
  FILTER(isIRI(?relation) && STRSTARTS(STR(?relation), 
         "http://rdf.freebase.com/ns/"))
} LIMIT 100

-- Incoming
SELECT DISTINCT ?relation WHERE {
  ?head ?relation ns:m.02mjmr .
  FILTER(isIRI(?relation) && STRSTARTS(STR(?relation),
         "http://rdf.freebase.com/ns/"))
} LIMIT 100
\end{lstlisting}

\subsection{\texttt{get\_triples} and CVT Flattening}
\label{sec:get_triples}

\texttt{get\_triples(entity, relations)} returns triples. For each relation, we query tails/heads using \texttt{GROUP BY} and \texttt{SAMPLE} to obtain one name per node. CVT nodes (no name or ID-only name) are flattened: we collect their relations, form two-hop names (e.g., \texttt{rel1.rel2}), and resolve via two-hop queries. Flatten relations are cached for future \texttt{get\_relations} calls. 
Details are shown in Listings~\ref{lst:triples_query}--\ref{lst:cvt_query}.

\begin{lstlisting}[caption={Triples query with aggregation.},label=lst:triples_query,float=ht]
PREFIX ns: <http://rdf.freebase.com/ns/>
SELECT ?tail (SAMPLE(?name_en) AS ?preferred_name)
             (SAMPLE(?name_any) AS ?fallback_name)
WHERE {
  ns:m.02mjmr ns:people.person.place_of_birth ?tail .
  OPTIONAL { 
    ?tail ns:type.object.name ?name_en .
    FILTER(LANGMATCHES(LANG(?name_en), 'en'))
  }
  OPTIONAL { ?tail ns:type.object.name ?name_any . }
}
GROUP BY ?tail LIMIT 40
\end{lstlisting}

\begin{lstlisting}[caption={CVT relation query.},label=lst:cvt_query,float=ht]
PREFIX ns: <http://rdf.freebase.com/ns/>
SELECT DISTINCT ?relation WHERE {
  ns:<cvt_node_id> ?relation ?tail .
  FILTER(isIRI(?relation) && STRSTARTS(STR(?relation),
         "http://rdf.freebase.com/ns/"))
} LIMIT 50
\end{lstlisting}

\newpage
\section{Training Implementation Details}
\label{app:training_details}

For the SFT stage, our implementation is based on the LlamaFactory framework \citep{zheng2024llamafactoryunifiedefficientfinetuning}; we use a cosine learning rate schedule with an initial learning rate of 1e-5, a batch size of 32, and a maximum sequence length of
4096. We fine-tune the model for 2 epochs. 
Reinforcement learning is implemented using the GRPO algorithm with the framework \emph{slime}; we set the number of rollouts to N=8, use a batch size of 64 for rollout, and a global batch size of 512 for training. We randomly sample a 6.4k subset from CWQ and train the model for 100 steps. 
All experiments were conducted on 8 × NVIDIA A100 (80GB) GPUs.
All hyperparameters and training configurations are reported in Table~\ref{tab:training-params}.

\begin{table}[ht]
\centering
\begin{tabular}{l c}
\toprule
\textbf{Parameter} & \textbf{Value} \\
\midrule
\multicolumn{2}{l}{\textbf{SFT settings}} \\
\midrule
Batch size & 32 \\
Learning Rate & 1e-5 \\
Max Response Length & 4096 \\
Epoch & 2 \\
\midrule
\multicolumn{2}{l}{\textbf{RL settings}} \\
\midrule
Rollout Batch Size & 64 \\
Rollout N & 8 \\
Global-batch Size & 512 \\
Learning Rate & 1e-6 \\
Max Prompt Length & 4096 \\
Max Response Length & 8192 \\
KL Loss Coefficient & 0.001 \\
Algorithm & GRPO \\
Steps & 100 \\
Rollout Temperature & 1.0 \\
\bottomrule
\end{tabular}
\caption{Key parameters during the training process.}
\label{tab:training-params}
\end{table}

\section{Prompts}
\label{app:prompts}

In this subsection, we list the prompts used for KG interaction, trajectory generation, question generation, data quality evaluation, and relation reranking. When using them, we replace the contents in \{\} with our target data.

%-----------------------------------------------------------------------------
% Box 1: KG Interaction Prompt for Evaluation
%-----------------------------------------------------------------------------
\begin{tcolorbox}[
  colback=white,
  colframe=blue!60,
  colbacktitle=blue!30,
  coltitle=white,
  fonttitle=\bfseries\scriptsize,
  fontupper=\scriptsize,
  title=KG Interaction Prompt,
  rounded corners,
  boxrule=0.5pt
]
You are a helpful assistant that answers questions based on knowledge graphs. You can query from knowledge base provided to you to answer the question up to \textcolor{teal!70!black}{\textit{\{max\_calls\}}} times.

\textbf{General Instructions:}
\begin{enumerate}
  \item First, perform careful reasoning inside \textcolor{blue}{\textit{<think>}}\textit{...}\textcolor{blue}{\textit{</think>}} tags. In this section, briefly outline your plan, record any working memory or intermediate observations, check for mistakes or ambiguous interpretations.
  \item If you need to query knowledge for more information, you can set a query statement between \textcolor{orange}{\textit{<kg-query>}}\textit{...}\textcolor{orange}{\textit{</kg-query>}} to query from knowledge graph. The query tool-using rules are provided in "KG Query Server Instruction" part. Each \textcolor{orange}{\textit{<kg-query>}} call must occur after \textcolor{blue}{\textit{<think>}}\textit{...}\textcolor{blue}{\textit{</think>}}, and you must not issue another \textcolor{orange}{\textit{<kg-query>}} until you have received the environment's \textcolor{purple}{\textit{<information>}} feedback for the previous query.
  \item If you have already found the answer, return it directly in the form \textcolor{red}{\textit{<answer>}}\textit{...}\textcolor{red}{\textit{</answer>}} and end your search. Your answer must strictly follow the KG Query Server Instruction.
\end{enumerate}

\textbf{KG Instructions:}

\textcolor{teal!70!black}{\textit{\{kg\_instruction\}}}

\textbf{Question:} \textcolor{teal!70!black}{\textit{\{question\}}}

Begin your search with the following initial entities. Use \textit{get\_relations} to explore their relations.

\textbf{Initial Entities:} \textcolor{teal!70!black}{\textit{\{initial\_entities\}}}
\end{tcolorbox}

%-----------------------------------------------------------------------------
% Box 2: KG Query Server Instructions (Shared by multiple prompts)
%-----------------------------------------------------------------------------
\begin{tcolorbox}[
  colback=white,
  colframe=blue!60,
  colbacktitle=blue!30,
  coltitle=white,
  fonttitle=\bfseries\scriptsize,
  fontupper=\scriptsize,
  title=KG Query Server Instructions,
  rounded corners,
  boxrule=0.5pt
]
\textbf{Available Query Functions:}
\begin{itemize}
  \item \textit{get\_relations(``entity\_name'')}: Returns all relations (incoming and outgoing) for the entity. Use this to explore the entity's properties.
  \item \textit{get\_triples(``entity\_name'', [``relation1'', ``relation2'', ...])}: Returns triples for the entity limited to the specified relations.
\end{itemize}

\textbf{KG Query Server Instructions:}
\begin{enumerate}
  \item If you encounter a KG-related error, read the error message carefully and correct your query.
  \item Use only the two query functions above; do not write natural-language queries inside \textcolor{orange}{\textit{<kg-query>}} tags.
  \item If Initial Entities are provided, begin your search from them using \textit{get\_relations(``initial\_entity'')}. If multiple initial entities are given, start with the most specific one. Analyze each systematically.
  \item Use the ENTITY NAME EXACTLY as shown in the \textcolor{purple}{\textit{<information>}} section. Example: \textit{get\_relations(``Barack Obama'')}.
  \item Before calling \textit{get\_triples(...)}, you MUST have called \textit{get\_relations(...)} for that entity (or already seen its relations) to ensure the relations exist.
  \item When calling \textit{get\_triples}, provide a list of ALL candidate relations returned by the previous step, ranked by relevance to the question. We will automatically use the top 4. Copy each relation EXACTLY as it appears in the \textcolor{purple}{\textit{<information>}} list. Example: \textit{get\_triples(``Barack Obama'', [``people.person.place\_of\_birth'', ``people.person.nationality'', ...])}.
  \item Always use double quotes around all function parameters (entity names and relations).
\end{enumerate}

\textbf{Answer Output Rules:}
\begin{enumerate}
  \item Format: \textcolor{red}{\textit{<answer>}}\textit{[``Answer1'', ``Answer2'', ...]}\textcolor{red}{\textit{</answer>}}. Return a JSON list of strings. The final answer must be a human-readable entity name, not a MID (e.g., m.01234). If the only candidate you have is a MID, do NOT output it as the final answer. Instead, explore the MID's neighbors (use \textit{get\_relations}/\textit{get\_triples}) to find a named entity to return.
  \item No external knowledge: answer ONLY using KG query results you have retrieved. The answer(s) MUST be EXACT and COMPLETE entity names from the retrieved triples. If multiple entities satisfy the question, list them all in the JSON list in \textcolor{red}{\textit{<answer>}} tags.
  \item Do NOT include any explanations in the \textcolor{red}{\textit{<answer>}} tags.
  \item NEVER include ``I don't know'' or ``Information not available'' in \textcolor{red}{\textit{<answer>}} tags. Provide the best possible answer(s) from the entities you have retrieved.
  \item CRITICAL: Your final answer must only contain entities that exist in the provided graph triples.
\end{enumerate}
\end{tcolorbox}

%-----------------------------------------------------------------------------
% Box 3: Trajectory Generation Prompt - Composition Type
%-----------------------------------------------------------------------------
\begin{tcolorbox}[
  colback=white,
  colframe=green!60,
  colbacktitle=green!30,
  coltitle=white,
  fonttitle=\bfseries\scriptsize,
  fontupper=\scriptsize,
  title=Trajectory Generation Prompt (Composition Type),
  rounded corners,
  boxrule=0.5pt
]
=== OVERVIEW ===

You are a helpful assistant that generates Supervised Fine-Tuning samples for a Knowledge Graph question-answering task. Your task is to generate a concise reasoning thought that explains WHY a specific action was chosen at a given step, based on the information you have gathered. Keep it brief and direct, following the style of the example provided.

=== AGENT ROLLOUT RULES ===

These rules describe how the underlying agent behaves. Your thoughts should align with them.

- Tools:
  - \textit{get\_relations(``entity\_name'')}: Returns all relations (both incoming and outgoing) connected to the entity.
  - \textit{get\_triples(``entity\_name'', [``relation1'', ``relation2'', ...])}: Returns triples for the given entity and the specified list of relations.

Entity reference rules (CRITICAL):
\begin{enumerate}
  \item Always use ENTITY NAME in all function calls. Copy names exactly as shown in the last \textcolor{purple}{\textit{<information>}}.
  \item If Initial entities are provided, you must start your search from them using \textit{get\_relations(``your\_initial\_entity'')}.
  \item Use double quotes around all function parameters (entity and relations).
  \item When calling \textit{get\_triples}, provide a list of relations that are helpful to answer the question, triples related to these relations will be automatically retrieved. Each relation must be copied EXACTLY from the latest \textcolor{purple}{\textit{<information>}} list. Example: \textit{get\_triples(``Barack Obama'', [``people.person.place\_of\_birth'', ``people.person.nationality''])}.
  \item Only use the two query functions listed above; do not write natural language queries inside \textcolor{orange}{\textit{<kg-query>}} tags.
  \item Before calling \textit{get\_triples(...)}, first call \textit{get\_relations(...)} for your current entity to obtain the predicate list.
  \item No outside knowledge: answer ONLY using KG results you have retrieved.
\end{enumerate}

=== WHAT TO WRITE ===

You will be given:
\begin{enumerate}
  \item The question to answer
  \item The topic entities (starting points)
  \item Historical information: all previous steps with their thoughts, actions, and observations
  \item The next action that should be taken (this is the "golden action" that leads to the answer)
\end{enumerate}

Your task is to write a concise thought that:
\begin{enumerate}
  \item Briefly mentions what you learned from the last observation (relations or triples)
  \item Directly explains why the given next action is the right choice at this point
\end{enumerate}

For the final step (when the next action is to provide an \textcolor{red}{\textit{<answer>}}):
\begin{itemize}
  \item Briefly explain how the observations from the last step provide the answer
\end{itemize}

Keep your thought concise and direct. Avoid over-summarizing or repeating information. Focus on the essential reasoning that leads to the action, similar to the example style.

=== CRITICAL CONSTRAINTS ===
\begin{enumerate}
  \item ONLY reference information from previous steps (historical information provided to you). NEVER mention entities, relations, or facts that may appear in future steps.
  \item Keep your thought concise and direct - typically 2-5 sentences, similar to the example style. Avoid lengthy summaries or over-explanation.
  \item Copy entity/predicate strings EXACTLY as shown in observations.
  \item Use natural language only; never include \textcolor{orange}{\textit{<kg-query>}} or \textcolor{red}{\textit{<answer>}} tags in your thoughts.
  \item Do NOT simply restate the action - explain the REASONING behind choosing it, but keep it brief.
\end{enumerate}

=== OUTPUT FORMAT ===

\textcolor{blue}{\textit{<think>}}

Your concise reasoning thought (typically 2-5 sentences, following the example's brief style)

\textcolor{blue}{\textit{</think>}}
\end{tcolorbox}

%-----------------------------------------------------------------------------
% Box 4: Trajectory Generation Prompt - Conjunction Type
%-----------------------------------------------------------------------------
\begin{tcolorbox}[
  colback=white,
  colframe=orange!60,
  colbacktitle=orange!30,
  coltitle=white,
  fonttitle=\bfseries\scriptsize,
  fontupper=\scriptsize,
  title=Trajectory Generation Prompt (Conjunction Type),
  rounded corners,
  boxrule=0.5pt
]
=== OVERVIEW ===

You are a helpful assistant that generates Supervised Fine-Tuning samples for a Knowledge Graph question-answering task involving CONJUNCTION reasoning. Your task is to generate a detailed reasoning thought that explains WHY a specific action was chosen at a given step, based on the historical information available up to that point.

CONJUNCTION reasoning means the question requires exploring from more than one different starting entities (topic entities) that eventually converge to the same answer entity through different paths. The agent is expected to explore all paths to find the common answer.

=== AGENT ROLLOUT RULES ===

These rules describe how the underlying agent behaves. Your thoughts should align with them.

- Tools:
  - \textit{get\_relations(``entity\_name'')}: Returns all relations (both incoming and outgoing) connected to the entity.
  - \textit{get\_triples(``entity\_name'', [``relation1'', ``relation2'', ...])}: Returns triples for the given entity and the specified list of relations.

Entity reference rules (CRITICAL):
\begin{enumerate}
  \item Always use ENTITY NAME in all function calls. Copy names exactly as shown in the last \textcolor{purple}{\textit{<information>}}.
  \item If Initial entities are provided, you must start your search from them using \textit{get\_relations(``your\_initial\_entity'')}.
  \item Use double quotes around all function parameters (entity and relations).
  \item When calling \textit{get\_triples}, provide a list of relations that are helpful to answer the question, triples related to these relations will be automatically retrieved. Each relation must be copied EXACTLY from the latest \textcolor{purple}{\textit{<information>}} list. Example: \textit{get\_triples(``Barack Obama'', [``people.person.place\_of\_birth'', ``people.person.nationality''])}.
  \item Only use the two query functions listed above; do not write natural language queries inside \textcolor{orange}{\textit{<kg-query>}} tags.
  \item Before calling \textit{get\_triples(...)}, first call \textit{get\_relations(...)} for your current entity to obtain the predicate list.
  \item No outside knowledge: answer ONLY using KG results you have retrieved.
\end{enumerate}

=== WHAT TO WRITE ===

You will be given:
\begin{enumerate}
  \item The question to answer
  \item The topic entities (starting points - may be multiple for conjunction questions)
  \item Historical information: all previous steps with their thoughts, actions, and observations
  \item The next action that should be taken (this is the "golden action" that leads to the answer)
\end{enumerate}

Your task is to write a concise thought that:
\begin{enumerate}
  \item Briefly mentions what you learned from the last observation (relations or triples)
  \item Directly explains why the given next action is the right choice at this point
\end{enumerate}

For the final step (when the next action is to provide an \textcolor{red}{\textit{<answer>}}):
\begin{itemize}
  \item Briefly explain how the observations from the previous step provide the answer
\end{itemize}

Keep your thought concise and direct. Avoid over-summarizing or repeating information. Focus on the essential reasoning that leads to the action, similar to the example style.

=== CRITICAL CONSTRAINTS ===
\begin{enumerate}
  \item ONLY reference information from previous steps (historical information provided to you). NEVER mention entities, relations, or facts that may appear in future steps.
  \item Keep your thought concise and direct - typically 2-5 sentences, similar to the example style. Avoid lengthy summaries or over-explanation.
  \item Copy entity/predicate strings EXACTLY as shown in observations.
  \item Use natural language only; never include \textcolor{orange}{\textit{<kg-query>}} or \textcolor{red}{\textit{<answer>}} tags in your thoughts.
  \item Do NOT simply restate the action - explain the REASONING behind choosing it, but keep it brief.
\end{enumerate}

=== OUTPUT FORMAT ===

\textcolor{blue}{\textit{<think>}}

Your concise reasoning thought (typically 2-5 sentences, following the example's brief style)

\textcolor{blue}{\textit{</think>}}
\end{tcolorbox}

%-----------------------------------------------------------------------------
% Box 5: Question Generation Prompts
%-----------------------------------------------------------------------------
\begin{tcolorbox}[
  colback=white,
  colframe=blue!60,
  colbacktitle=blue!30,
  coltitle=white,
  fonttitle=\bfseries\scriptsize,
  fontupper=\scriptsize,
  title=Question Generation Prompts,
  rounded corners,
  boxrule=0.5pt
]
\vspace{0.5em}
\hrule
\vspace{0.3em}

\textbf{Single-Path Question:}

You are given:

- A path represented as an ordered sequence of relations, topic entity and answer entity (e.g., Topic Entity --relationA--> entity1 --> relationB--> entity2 --> ... --> Answer Entity).

You will be provided with the name of the topic entity and the name of the answer entity. Names of intermediate entities in the path will be replaced by placeholders like entity1, entity2, etc.

\textbf{Your task:}
\begin{itemize}
  \item Compose ONE concise English question whose unique answer is exactly the Answer Entity.
  \item Implicitly reflect the multi-hop path, but keep it brief and natural. Avoid enumerations or multiple questions.
  \item STRICT: The question must contain exactly one question mark '?' (only one interrogative). The question must be a single sentence with only one '?'.
  \item Do NOT directly mention raw relation labels. Paraphrase using concrete, meaningful semantics (e.g., country, city, river, award, university, founder, member of, genre, located in, part of, spouse, parent, date, birthplace, capital).
  \item Keep the question concise (ideally < 100 characters).
  \item Use only the information implied by the path; do not introduce outside facts.
  \item Try to use different styles and sentence structures when generating questions.
\end{itemize}

\textbf{Input:}

Path: \textcolor{teal!70!black}{\textit{\{path\}}}

CRITICAL: Do NOT include any explanations, analysis, or additional text. Output ONLY the one line below, nothing more.

Output format (STRICT --- output EXACTLY one line, nothing else):

\textcolor{green!70!black}{\textit{<question>}}\textit{(one concise question with exactly one '?')}\textcolor{green!70!black}{\textit{</question>}}

\vspace{0.5em}
\hrule
\vspace{0.3em}

\textbf{Conjunction Question:}

You are given:
- Two paths that both lead to the same answer entity:

Your task:
\begin{itemize}
  \item Compose ONE complete and concise English question whose unique answer is the common entity reached by both paths (the final entity in both paths).
  \item The question should implicitly reflect that the answer requires exploring from BOTH starting entities (conjunction reasoning), not just one of them.
  \item Keep it brief and natural (ideally < 120 characters). Avoid enumerations or multiple questions.
  \item STRICT: The question must be COMPLETE and contain exactly one question mark '?' (only one interrogative). Prefer a single sentence (or two short clauses joined), but only one '?'.
  \item Do NOT directly mention exact entity names/IDs or raw relation labels. Paraphrase using concrete, meaningful semantics (e.g., country, city, river, award, university, founder, member of, genre, located in, part of, spouse, parent, date, birthplace, capital).
  \item De-emphasize meta/technical relations (e.g., types, notable\_for, generic topic) unless essential. Focus on relations with clear real-world meaning.
  \item Use only the information implied by the two paths; do not introduce outside facts.
  \item Try to use diverse styles and sentence structures when generating questions to avoid repetitive patterns and enhance question diversity.
  \item CRITICAL: Output a COMPLETE question. Do NOT truncate or cut off the question. The question must end with a question mark '?'.
  \item You MUST output the question in the format \textcolor{green!70!black}{\textit{<question>}}\textit{<your complete question>}\textcolor{green!70!black}{\textit{</question>}}
\end{itemize}

\textbf{Input:}

Path 1: \textcolor{teal!70!black}{\textit{\{path1\}}}

Path 2: \textcolor{teal!70!black}{\textit{\{path2\}}}

\textbf{Output:}

\textcolor{green!70!black}{\textit{<question>}}\textit{<your complete question>}\textcolor{green!70!black}{\textit{</question>}}
\end{tcolorbox}

%-----------------------------------------------------------------------------
% Box 6: Data Quality Evaluation Prompt Template
%-----------------------------------------------------------------------------
\begin{tcolorbox}[
  colback=white,
  colframe=blue!60,
  colbacktitle=blue!30,
  coltitle=white,
  fonttitle=\bfseries\scriptsize,
  fontupper=\scriptsize,
  title=Data Quality Evaluation Prompt Template,
  rounded corners,
  boxrule=0.5pt
]
You are an expert evaluator for Knowledge Graph Question Answering (KGQA) data quality. Your task is to evaluate question-path pairs based on multiple quality criteria.

\textbf{Input Format}

You will be provided with two inputs:
\begin{enumerate}
  \item \textbf{name\_path}: A reasoning path in the knowledge graph, represented as a sequence of entities and relations connected by ``->''
  \item \textbf{question}: The natural language question that should be answerable by following this path
\end{enumerate}

Input Example:

\textit{name\_path}: ``Emap International Limited -> organization.organization.founders -> Richard Winfrey -> people.person.place\_of\_birth -> Long Sutton, Lincolnshire''

\textit{question}: ``What is the birthplace of a founder of Emap International Limited?''

\textbf{Task}

For each question-path pair provided, you must:
Evaluate it on the following dimension and assign a score from 0 to 10 based on the scoring guide:

\textbf{Evaluation Criteria}

You must evaluate each question-path pair based on the following dimension:
\textcolor{teal!70!black}{\textit{\{dimension\_prompt\}}}

\textbf{Important Guidelines}
\begin{enumerate}
  \item \textbf{Strict Evaluation}: Be strict but fair.
  \item \textbf{Output Consistency}: Always output the score in the exact format above. Do not include any text outside the valid structure.
  \item Do not include any other text or explanation.
\end{enumerate}

\textbf{Output Format}

You must output only a single score from 0 to 10:
\textcolor{gray!60!black}{\textit{<score>}}\textit{<0-10>}\textcolor{gray!60!black}{\textit{</score>}}
\end{tcolorbox}

%-----------------------------------------------------------------------------
% Box 7: LLM Relation Reranking Prompt
%-----------------------------------------------------------------------------

\begin{tcolorbox}[
  colback=white,
  colframe=blue!60,
  colbacktitle=blue!30,
  coltitle=white,
  fonttitle=\bfseries\scriptsize,
  fontupper=\scriptsize,
  title=Relation Reranking Prompt Template,
  rounded corners,
  boxrule=0.5pt
]
You are an expert Knowledge Graph navigator. Your goal is to select the most relevant relations that will lead to the answer for a given question.

\textbf{You will be provided with:}
\begin{enumerate}
    \setlength{\itemsep}{0pt}
    \setlength{\parskip}{0pt}
    \item A Question.
    \item A Topic Entity involved in the question.
    \item A list of Candidate Relations connected to that entity.
    \item A Previous Model Thinking Process: The reasoning from the model's last turn (if available), which may provide helpful context about the query strategy and what information is being sought.
\end{enumerate}

\textbf{Task (STRICT RULES):}
\begin{itemize}
    \setlength{\itemsep}{0pt}
    \setlength{\parskip}{0pt}
    \item You MUST select relations ONLY from the provided Candidate Relations list.
    \begin{itemize}
        \item Each selected string MUST be an EXACT character-by-character match to an item in the input list.
        \item Do NOT rename relations, do NOT change underscores/dots, do NOT add/remove numeric suffixes.
    \end{itemize}
    \item You MUST NOT invent new relations.
    \item You MUST NOT output an empty list if at least one candidate relation is provided.
    \item Output length rules (use K = \{k\}, which matches the evaluation setting \texttt{kg\_top\_k}):
    \begin{itemize}
        \item If the input list has at least \{k\} candidates, return EXACTLY \{k\} DISTINCT relations.
        \item If the input list has fewer than \{k\} candidates, return ALL candidates (still non-empty), and do NOT pad with invented items.
    \end{itemize}
    \item Rank them by relevance (most relevant first).
    \item Output ONLY a JSON array of strings. Do not include any other text.
\end{itemize}

\textbf{Examples} (illustrative format; when K differs from the example length, still follow the rules above with your K):

Question: \{\} \\
Topic Entity: \{\} \\
Relations: \{\} \\
Your Selections: \{\} 
\end{tcolorbox}

%-----------------------------------------------------------------------------
% Box 8: Evaluation Dimension Prompts (Three dimensions in one box)
%-----------------------------------------------------------------------------
\vspace{0.3em}

\begin{tcolorbox}[
  colback=white,
  colframe=blue!60,
  colbacktitle=blue!30,
  coltitle=white,
  fonttitle=\bfseries\scriptsize,
  fontupper=\scriptsize,
  title=Evaluation Dimension Prompts,
  rounded corners,
  boxrule=0.5pt
]
\textbf{Reasoning Path Quality (0--10)}

Reasoning Path Quality: ``Whether the reasoning path forms a complete, valid, and efficient chain from the starting entity to the answer entity, with all relations semantically correct and entities properly resolved.''

\textbf{Scoring Guide (0--10)}
\begin{itemize}
  \item \textbf{10}: The path is complete, direct, and optimal. All relations are semantically valid and correctly connected. All entities are properly resolved with no ambiguity. The path represents the most efficient route to the answer with no unnecessary steps.
  \item \textbf{8--9}: The path is complete and valid, with only minor inefficiencies or one minor ambiguity. All relations are correct, and the path successfully leads to the answer entity.
  \item \textbf{6--7}: The path is mostly complete and valid, but may have some inefficiencies, minor relation issues, or one ambiguous entity. The path can still lead to the answer, though not optimally.
  \item \textbf{4--5}: The path has noticeable issues: missing intermediate entities, some invalid relations, or multiple ambiguous entities. The path may still reach an answer but with significant problems.
  \item \textbf{1--3}: The path has serious structural problems: circular references, multiple invalid relations, or critical missing entities. The path may not reliably lead to the answer.
  \item \textbf{0}: The path is fundamentally broken: cannot form a valid chain, contains only invalid relations, or fails to connect to any meaningful answer entity.
\end{itemize}

\vspace{0.4em}
\hrule
\vspace{0.4em}

\textbf{Question-Path Relevance (0--10)}

Question-Path Relevance: ``Whether the question semantically aligns with what the path can answer, and whether the path contains all necessary information to fully and accurately answer the question.''

\textbf{Scoring Guide (0--10)}
\begin{itemize}
  \item \textbf{10}: Perfect alignment. The question directly matches what the path answers. The path contains all necessary information and fully addresses the question. The answer entity at the end of the path is exactly what the question asks for.
  \item \textbf{8--9}: Strong alignment with minor gaps. The path answers the question well, with only very minor information missing or slight semantic nuances not perfectly captured.
  \item \textbf{6--7}: Good alignment but incomplete. The path addresses the core of the question but may miss some aspects or provide partial information. The answer entity is relevant but may not fully satisfy the question.
  \item \textbf{4--5}: Moderate alignment with notable gaps. The path is related to the question but doesn't fully answer it, or answers a different but related aspect. Some necessary information is missing.
  \item \textbf{1--3}: Poor alignment. The path is tangentially related but doesn't answer the question, or answers a different question entirely. Significant information mismatch.
  \item \textbf{0}: No alignment. The path and question are completely unrelated, or the path cannot answer the question at all.
\end{itemize}

\vspace{0.4em}
\hrule
\vspace{0.4em}

\textbf{Question Semantic Coherence (0--10)}

Question Semantic Coherence: ``Whether the question is grammatically correct, semantically clear, naturally phrased, and logically structured like a human would ask it.''

\textbf{Scoring Guide (0--10)}
\begin{itemize}
  \item \textbf{10}: Perfect coherence. The question is grammatically flawless, semantically crystal clear, reads naturally, and has impeccable logical structure. It sounds exactly like a native speaker would ask it.
  \item \textbf{8--9}: Excellent coherence with minor imperfections. The question is grammatically correct, clear, and natural, with only very minor stylistic issues or slight ambiguity.
  \item \textbf{6--7}: Good coherence with some issues. The question is mostly grammatically correct and understandable, but may have minor grammatical errors, slight ambiguity, or somewhat unnatural phrasing.
  \item \textbf{4--5}: Moderate coherence with noticeable problems. The question has grammatical errors, some ambiguity, or unnatural phrasing that affects comprehension, but the core meaning is still discernible.
  \item \textbf{1--3}: Poor coherence. The question has significant grammatical errors, is confusing or ambiguous, or has very unnatural phrasing that makes it difficult to understand.
  \item \textbf{0}: No coherence. The question is grammatically broken, completely ambiguous, or makes no logical sense. It cannot be understood as a valid question.
\end{itemize}
\end{tcolorbox}

\end{document}